\documentclass[11pt]{article}

\usepackage{etoolbox}

\usepackage{multirow}
\usepackage{tabularx}
\usepackage{longtable}

\begingroup 
    \catcode `\@ = 11
    \catcode `\~ = 13
    \catcode `\% = 12
    \protected\long\gdef\cmt@remove#1
    \ifdefined~
        \global\let\cmt@old~
    \else
        \global\let\cmt@old\relax
    \fi
    \protected\gdef~{\begingroup\catcode`
        \futurelet\next\cmt@}
    \protected\gdef\cmt@
      {\ifx
           \expandafter\cmt@remove
       \else
           \endgroup\expandafter\cmt@old
       \fi}
\endgroup
   ~%
       Lines
       to
       comment

\usepackage{cite}

    \usepackage{microtype} 

    \usepackage{pgf} 

    \usepackage{float} 

    \usepackage{graphicx}
    \graphicspath{{./images/}}

    \usepackage{caption}
  
    \usepackage{subcaption}
\usepackage{booktabs,mathptmx,siunitx} 
\usepackage{hhline} 
\usepackage{mathtools}
\usepackage{amsfonts,amssymb}
\usepackage{wasysym}  

\usepackage{xifthen}
\usepackage{colonequals} 

\usepackage[algoruled,vlined]{algorithm2e}
\usepackage{url}
\usepackage{enumerate}
\usepackage{bm}
\usepackage{tabularx}
\usepackage{flushend}
\usepackage{hyperref}
\usepackage[plain]{fancyref}  
~%
\usepackage{type1cm,eso-pic,color}
\makeatletter
          \AddToShipoutPicture{
            \setlength{\@tempdimb}{.5\paperwidth}
            \setlength{\@tempdimc}{.5\paperheight}
            \setlength{\unitlength}{1pt}
            \put(\strip@pt\@tempdimb,\strip@pt\@tempdimc){
        \makebox(0,0){\rotatebox{55}{\textcolor[gray]{0.85}
        {\fontsize{5cm}{5cm}\selectfont{DRAFT}}}}
            }
        }
\makeatother

\usepackage{cancel}

\usepackage{commath}

\DeclarePairedDelimiter\myabs{\lvert}{\rvert}%
\DeclarePairedDelimiter\mynorm{\lVert}{\rVert}%

\makeatletter
\let\oldabs\myabs
\def\myabs{\@ifstar{\oldabs}{\oldabs*}}
\let\oldnorm\mynorm
\def\mynorm{\@ifstar{\oldnorm}{\oldnorm*}}
\makeatother

\DeclareMathOperator*{\argmax}{\arg\,max}

\newcommand{\asin}[1]{\arcsin \negmedspace \left( #1 \right)}
\newcommand{\acos}[1]{\arccos \negmedspace \left( #1 \right)}
\newcommand{\atan}[1]{\arctan \negmedspace \left( #1 \right)}

\newcommand{\point}[1]{\mathrm{\uppercase{#1}}}

\newcommand{\pixelPoint}[1]{\boldsymbol{#1}}

\newcommand{\func}[1]{\mathrm{\lowercase{#1}}}
\newcommand{\funcvv}[1]{\mathbf{\lowercase{#1}}}

\newcommand{\trans}[1]{{#1}^{\mathsf{T}}}

\newcommand{\vect}[1]{\mathbf{\lowercase{#1}}}

\newcommand{\matx}[1]{\mathbf{\uppercase{#1}}}

\newcommand{\fr}[1]{%
    \hbox{\ensuremath{
    \scalebox{0.9}{$\left[\mathrm{#1}\right]$}
    }}
}

\newcommand{\cfs}[3]{
    \hbox{\ensuremath{
    ^{
      \ifthenelse{\isempty{#3}}%
        {}
        {\scalebox{0.6}{\fr{#3}}}
    }
    _{
        \ifthenelse{\isempty{#2}}%
        {}
        {\scalebox{0.6}{\fr{#2}}}
    }
    {\negthinspace #1}
    }}
}

\newcommand{\vectThreeRow}[3]{
    \hbox{\ensuremath{
    \begin{bmatrix}#1 \\ #2 \\ #3\end{bmatrix}
    }}
    }

\newcommand{\vectTwoCol}[2]{
 [\begin{matrix}{#1},&{#2}\end{matrix}]
    }
\newcommand{\vectThreeCol}[3]{
 [\begin{matrix}{#1},&{#2},&{#3}\end{matrix}]
    }
\newcommand{\vectFourCol}[4]{
 [\begin{matrix}{#1},&{#2},&{#3},&{#4}\end{matrix}]
    }

\newcommand{\vectTwoColInline}[2]{
 \trans{\vectTwoCol{#1}{#2}}}
\newcommand{\vectThreeColInline}[3]{
 \trans{\vectThreeCol{#1}{#2}{#3}}}
\newcommand{\vectFourColInline}[4]{
 \trans{\vectFourCol{#1}{#2}{#3}{#4}}}

\DeclareMathSymbol{\Gamma}{\mathalpha}{operators}{0}
\DeclareMathSymbol{\Delta}{\mathalpha}{operators}{1}
\DeclareMathSymbol{\Lambda}{\mathalpha}{operators}{3}
\DeclareMathSymbol{\Xi}{\mathalpha}{operators}{4}
\DeclareMathSymbol{\Pi}{\mathalpha}{operators}{5}
\DeclareMathSymbol{\Sigma}{\mathalpha}{operators}{6}
\DeclareMathSymbol{\Upsilon}{\mathalpha}{operators}{7}
\DeclareMathSymbol{\Phi}{\mathalpha}{operators}{8}
\DeclareMathSymbol{\Psi}{\mathalpha}{operators}{9}
\DeclareMathSymbol{\Omega}{\mathalpha}{operators}{10}

\usepackage{upgreek}
\newcommand{\vecttheta}{\pmb{\uptheta}}
\newcommand{\vectmu}{\pmb{\upmu}}

~
\makeatletter
\newlength \figwidth
\if@twocolumn
\else
\fi
\makeatother
\includegraphics[width=\figwidth]{myimage}

\usepackage{tikz}
\tikzset{
    MyPersp/.style={scale=1.8,x={(-0.8cm,-0.4cm)},y={(0.8cm,-0.4cm)},
    z={(0cm,1cm)}},
    MyPoints/.style={fill=white,draw=black,thick}
        }

\newcommand*\mycirc[1]{%
  \begin{tikzpicture}[baseline=(C.base)]
    \node[draw,circle,inner sep=1pt](C) {#1};
  \end{tikzpicture}}

\numberwithin{equation}{section}

\usepackage{color}
\definecolor{gray}{rgb}{0.5,0.5,0.5}

\begin{document}
%
\author{Carlos Jaramillo\\e$-$mail: {\tt \small cjaramillo@gradcenter.cuny.edu}} 

\title{Design and Analysis of a Single$-$Camera Omnistereo Sensor for Quadrotor Micro Aerial Vehicles (MAVs)}


\maketitle

\begin{abstract}
  
We describe the design and 3D sensing performance of an omnidirectional
stereo-vision system (omnistereo) as applied to Micro Aerial Vehicles (MAVs). 
The proposed omnistereo model employs a monocular camera that is co-axially aligned with a
pair of hyperboloidal mirrors (folded catadioptric configuration). We show that this arrangement is
practical for performing stereo-vision when mounted on top of propeller-based MAVs characterized by low payloads.
The theoretical single viewpoint (SVP) constraint helps us derive analytical
solutions for the sensor's projective geometry and generate
SVP-compliant panoramic images to compute 3D information from stereo correspondences (in
a truly synchronous fashion). We perform an extensive analysis on various system
characteristics such as its size, catadioptric spatial resolution, field-of-view. In addition, we
pose a probabilistic model for uncertainty estimation of the depth from triangulation for skew
back-projection rays.
We expect to motivate the reproducibility of our
solution since it can be adapted (optimally) to other catadioptric-based omnistereo vision
applications.
\end{abstract}


\section{Introduction}
\label{sec:Introduction}

\subsection{Motivation}
\label{sec:motivation}
Micro aerial vehicles (MAVs), such as quadrotor helicopters, are
popular platforms for unmanned aerial vehicle (UAV) research due to their
structural simplicity, small form factor, their vertical take-off and landing
(VTOL) capability, and high omnidirectional maneuverability.
In general, UAVs have plenty of military and civilian applications, such as target localization
and tracking, 3-dimensional (3D) mapping, terrain and infrastructural inspection,
disaster monitoring, environmental and traffic surveillance, search and rescue, 
deployment of instrumentation, and cinematography, among other uses. However, with MAVs, their
payload and on-board computation limitations constrain their sensors to be compact and lightweight
and to execute efficient signal-processing algorithms with data obtained from as few sensors
as possible to avoid synchronization delays.
The most commonly used perception sensors on MAVs are laser scanners and
camera(s) in various configurations such as monocular, stereo, or omnidirectional.
Lightweight 2.5D laser scanners like those made by
Hokuyo\textsuperscript{\textregistered} exist and can accurately measure
distances at fast rates, however, they are limited to plane sweeps, which in turn
require the quadrotor to move up and down constantly in order to generate full 3D
maps or to foresee obstacles and free space for navigation. More recently, 3D laser/lidar
rangefinders have started to emerge, but their size is still an issue for MAVs applications. One
last disadvantage of laser rangefinders is their active sensing nature, which consumes more power
than cameras and are vulnerable to signal
corruption (e.g.
due dark/reflective surfaces).
Red, green, blue plus depth (RGB-D) sensors like the Microsoft
Kinect\textsuperscript{\textregistered} are also very popular for 3D navigation, but they have been
adopted for mainly indoor navigation \cite{Valenti2014} due to its structured infrared light
projection and short range sensing under \SI{5}{\metre}.
Hence, a lightweight imaging system of compact structure and capable of
providing a large field of view (FOV) instantaneously at acceptable resolutions is necessary
for MAV applications in 3D space. These requirements motivate the design and analysis of our
omnidirectional stereo (omnistereo) sensor detailed in this manuscript.

\subsection{Related Work}
\label{sec:related_work}
In general, stereo vision is used to produce dense or sparse
depth maps of the overlapping view instantenously, and its various uses are well known, for example,
in 3D reconstruction, in generating occupancy grids for obstacle avoidance, etc., including its
application to AUVs \cite{Byrne2006}.
Omnidirectional stereo based on multi-view camera arrangements have also been formulated through
the literature. Alternatively, omnidirectional `catadioptric' vision
systems are a possible solution employing cameras and mirrors \cite{Gluckman1998a}.
Originally, Nayar and Peri \cite{Nayar1999} studied 9 possible folded configurations for a
single-camera omnistereo imaging system.
Throughout the years, 
\cite{Koyasu2001}\cite{Bajcsy2003}\cite{Cabral2004a}\cite{Su2005}\cite{Mouaddib2005}\cite{Schonbein2011}
are some of the works that have implemented various of omnistereo catadioptric configurations for
ground mobile robots. 
Unfortunately, these systems are not compact since they use separate camera-mirror pairs, which 
it is known to cause synchronization issues. In fact, omnidirectional vision using a single mirror for flying of large UAVs
was first attempted in \cite{Hrabar2003}. In \cite{Hrabar2008}, Hrabar proposed 
the use of traditional horizontal stereo-based obstacle avoidance and path planing for AUVs, but
these techniques were only tested in a scaled-down air vehicle simulator (AVS).
In fact, we believe we are the first to present a single-camera catadioptric
approach to omnistereo vision for MAVs. The initial geometry of our model was proposed in
\cite{Guo2010} and the first prototypes were fabricated in \cite{Jaramillo2013a}.

It is true that a omnidirectional catadioptric system sacrifices spatial resolution on the imaging
sensor (as we analyze in \Fref{sec:spatial_resolution}). Higher resolution panoramas could be
achieved by rotating a linear camera as presented in \cite{Smadja2006}, but this approach has
severe disadvantages in dynamic environments.
Hence, our sensor offers practical advantages such as reduced cost, weight, and truly-instantaneous
pixel-disparity correspondences since a single camera does not introduce extra discrepancies on its intrinsic parameters or mis-synchronization issues.
In our previous work \cite{Labutov2011}, we developed a novel omnidirectional omnistereo
catadioptric rig consisting of a perspective camera coaxially-aligned with two spherical mirrors of distinct radii (in a ``folded"
configuration). One caveat of spherical mirrors is their
non-centrality as these do not
satisfy the single effective viewpoint (SVP) constraint (discussed in \Fref{sec:SVP_configuration}).
Instead, a locus of viewpoints is obtained rather than the more desirable SVP projection.
\cite{Swaminathan2001}.
Therefore, we extend our research to design a SVP-compliant omnistereo system
based on a folded, catadioptric configuration with hyperboloidal mirrors.
Yi and Ahuja implemented a few configurations (mixing mirrors and lenses) for omnidirectional stereo
vision system using a single camera and capable of a lengthy baseline\cite{Yi2006}. In \cite{He2008}, He
et al. also implemented a folded catadioptric system with hyperbolic mirrors for wider baseline and
they proposed various solutions to the stereo matching problem for their panoramic images (something
not dealt with here, and we refer the reader to \cite{Arican2009}).
Indeed, our approach resembles the work of Jang, Kim, and Kweon in \cite{Jang2005}, who
also proved the concept for a single camera catadioptric stereo system using hyperbolic mirrors.
However, their sensor's characteristics were not analyzed in order to justify their design
parameters and capabilities. 
Actually, we perform an elaborate analysis of our model's parameters
(\Fref{sec:sensor_design}) involving its geometric projection (\Fref{sec:projective_geometry}) that are obtained as
a constrained numerical optimization solution devising the sensor's real-life
application to MAVs passive range sensing (\Fref{sec:optimization_and_prototyping}). 
We also show how the panoramic images are obtained, where we find correspondences and triangulate 3D
points for which an uncertainty model is introduced (\Fref{sec:3D_sensing}). We finally discuss our
experimental results for 3D sensing from omnistereo and discuss future directions of our work
(\Fref{sec:experiments_and_future_work}).

\begin{figure}
    \centering
    \includegraphics[height=0.5\textwidth,keepaspectratio]{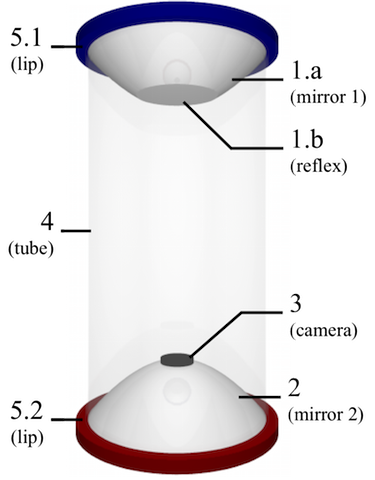}
    \caption[Synthetic model of the catadioptric omnistereo rig.]{Synthetic
    catadioptric single-camera omnistereo system, whose main tangible are
    components: 1.a-b) hyperboloid-planar combined mirror at top, 
    2) hyperboloidal mirror at bottom, 3) camera, 4) transparent cylindrical
    support, and 5.1-2) blue-red lips acting as circular markers
    during image processing. }
    \label{fig:sim_rig_alone}
\end{figure}        

\subsection{Symbol Notation}
\label{sec:notation}

In this subsection, we present the notation style used throughout the remaining of this
manuscript:

\begin{itemize}
  \item 
  \begin{par}
  With $\mathbb{R}^3$ as the set of all real points in Euclidean 3D space, a point $\point{P}_i \in
  \mathbb{R}^3$ uses a post-subscript $i$ to be identified from all the other points $P$, a
  strict subset of $\mathbb{R}^3$ or symbolically: $P \subset \mathbb{R}^3$ such that  $\point{P}_i \bigcup P =  \mathbb{R}^3$.
  \end{par}
  
  \item
  \begin{par}
  The origin of a coordinates system or frame is denoted by point
  $\point{O_A}$, where the subscript indicates the frame's name. In this case, $\fr{A}$ is the
  reference frame, which we put in square brackets (``$[~]$'') for clarity. We also
  name images with this frame notation style.
  \end{par}

  \item
  \begin{par}
  The position of a 3D point $\point{P}_i$ with respect to a reference frame $\fr{A}$ is indicated
  component-wise by a column vector of similar name (written in boldface upright lowercase) such as
  $\cfs{\vect{p}_i}{}{A} = \vectThreeColInline{x_i}{y_i}{z_i}$. Notice how the frame $\fr{A}$, which
  vector $\vect{p}_i$ is referred with respect to, is written as a pre-superscript on a symbol.
  Using homogeneous coordinates, the same
  vector is indicated with an $h$ post-subscript as in $\cfs{\vect{p}_{i,h}}{}{A} =
  \vectFourColInline{x_i}{y_i}{z_i}{1}$.
  \end{par}
    
  \item
  \begin{par}
  Differing from the 3D vector notation style, the position vector of a 2D point is
  written in boldface italic lowercase, such as $\cfs{\pixelPoint{m}_i}{}{I}$ for a pixel position
  on image frame $\fr{I}$.
  \end{par}
  
  \item
  \begin{par}
  As usual, the transpose of a vector or a matrix is indicated by a
  post-superscript ``$\mathsf{T}$'' symbol, such as $\trans{\vect{p}}$.
  \end{par} 
  
  \item
  \begin{par}
  Double vertical bars (``$\mid\mid$'') on each side of a vector expression are used to
  denote its magnitude (Euclidean norm). By definition, $\norm{\vect{p}_i} =
  \sqrt{{x_i}^2+{y_i}^2+{z_i}^2}$.
  \end{par}
  
  
  \item
  \begin{par}
  A unit vector wears a
  caret (``$\char`\^$'') on top. For instance, $\hat{\vect{q}}$ obeys $\norm{\hat{\vect{q}}} =
  1$.
  \end{par}

  \item
  \begin{par}
  Matrices are represented in boldface upright uppercase. A matrix $\matx{M}_i$
  will be written as $\matx{M}_{i,h}$ in homogeneous coordinates.
  \end{par}
  
  \item
  \begin{par}
  Function names use upright lowercase letters and additional subscript names based on
  context. To distinguish vector-valued functions from scalar-valued functions, we use
  boldface, such as for the backward projection function $\funcvv{f}_\beta$.
  \end{par}


\end{itemize}

\section{Sensor Design}
\label{sec:sensor_design}

\begin{figure} 
\centering
    \begin{subfigure}[c]{.5\textwidth}
            \includegraphics[width=\columnwidth,
            keepaspectratio]{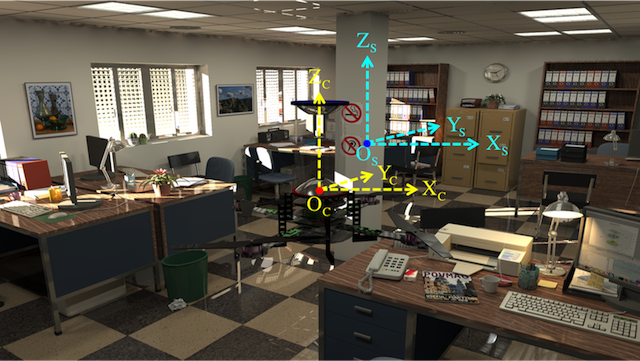}
            \subcaption{External view }
            \label{fig:sim_rig_world_view}
    \end{subfigure}
    \vfill
    \begin{subfigure}[c]{.5\textwidth}
            \includegraphics[width=\columnwidth,
            keepaspectratio]{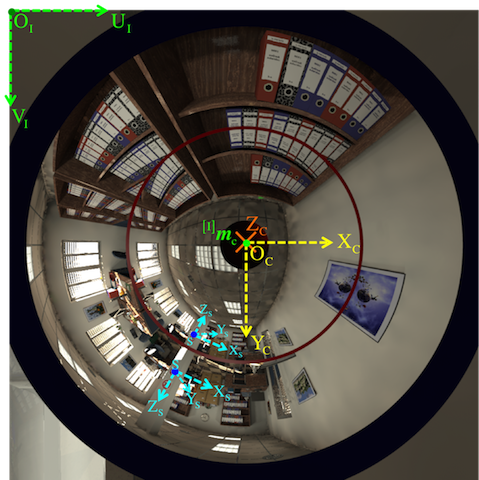}
            \subcaption{Camera view}
            \label{fig:sim_rig_camera_view}
    \end{subfigure}
    
    \caption[Synthetic scene.]{Photo-realistic synthetic scene:
            {\bf(\subref{fig:sim_rig_world_view})} Side-view of the quadrotor
            with the omnistereo rig in an office environment (the scene frame
            $\fr{S}$ and the system camera frame $\fr{C}$ can be appreciated).
            {\bf (\subref{fig:sim_rig_camera_view})} the image captured by 
            the system's camera using the same pose shown in
             \Fref{fig:sim_rig_world_view}'s scene. $\point{O_I}$ is the origin of the image
             frame $\fr{I}$ 
             and is located at the top-left corner, whereas the camera frame
             $\fr{C}$ coincides with the image's center point $\cfs{\pixelPoint{m}_c}{}{I}$. Here, 
             the camera's optical axis is coincident with the $+\point{Z_C}$-axis, which is
             perpendicularly directed into the virtual image plane $\pi$.
             Notice how objects reflected by the bottom mirror appear in the inner image ring bounded by a
             red circle mark, whereas reflections from the top mirror are imaged
             in the outer ring.
             Radial correspondences of pixels around the image's
            center facilitate the computation of depth explained in 
            (\Fref{sec:3D_sensing}).}
    \label{fig:sim_office}
\end{figure}

\Fref{fig:sim_rig_alone} shows the single-camera catadioptric omnistereo
vision system that we specifically design to be mounted on top of our
micro quadrotors (manufactured by Ascending Technologies \cite{AscTec2014}). It consists
of 1) one hyperboloid-planar mirror at the top, 2) one hyperboloidal
mirror at the bottom, and 3) a high-resolution USB camera also at the
bottom (inside the bottom mirror and looking up). The components are housed and supported by a 4) 
transparent tube or plastic standoffs (for the real-life prototype shown in
\Fref{fig:prototyped_rigs}).
The circular lips (5.1-2) that extend out of each mirror are respectively colored blue and red and serve to assist with the
boundary recognition during calibration (not discussed in this work).
The choice of the hyperboloidal reflectors owes to three reasons: it is one of
the four non-degenerated conic shapes satisfying the SVP constraint
\cite{Baker1999}; it allows a wider vertical FOV than elliptical and planar
mirrors; and it does not require a telescopic (orthographic) lens for imaging as
with paraboloidal mirrors (so our system can be downsized).
In addition, the planar part of mirror 1 works as a reflex mirror, which in part
reduces distortion caused by dual conic reflections.
Based on the SVP property, the system obtains two radial images of the
omnidirectional views in the form of an inner and an outer ring as illustrated
in \Fref{fig:sim_rig_world_view} and \Fref{fig:sim_rig_camera_view}).
Nevertheless, the unique set of parameters describing the entire system
categorizes it as a ``global camera model" given by \cite{Sturm2011a} because
changing the value of any parameter in the model affects the overall projection
function of visible light rays in the scene (imaging) as well as other
computational imaging factors such as depth resolution and overlapping field of
view, which we attempt to optimize with the following design subsections.

\subsection{Model Parameters}
\label{sec:model_parameters}
\begin{figure}
    \centering
    \includegraphics[width=0.7\textwidth,keepaspectratio]{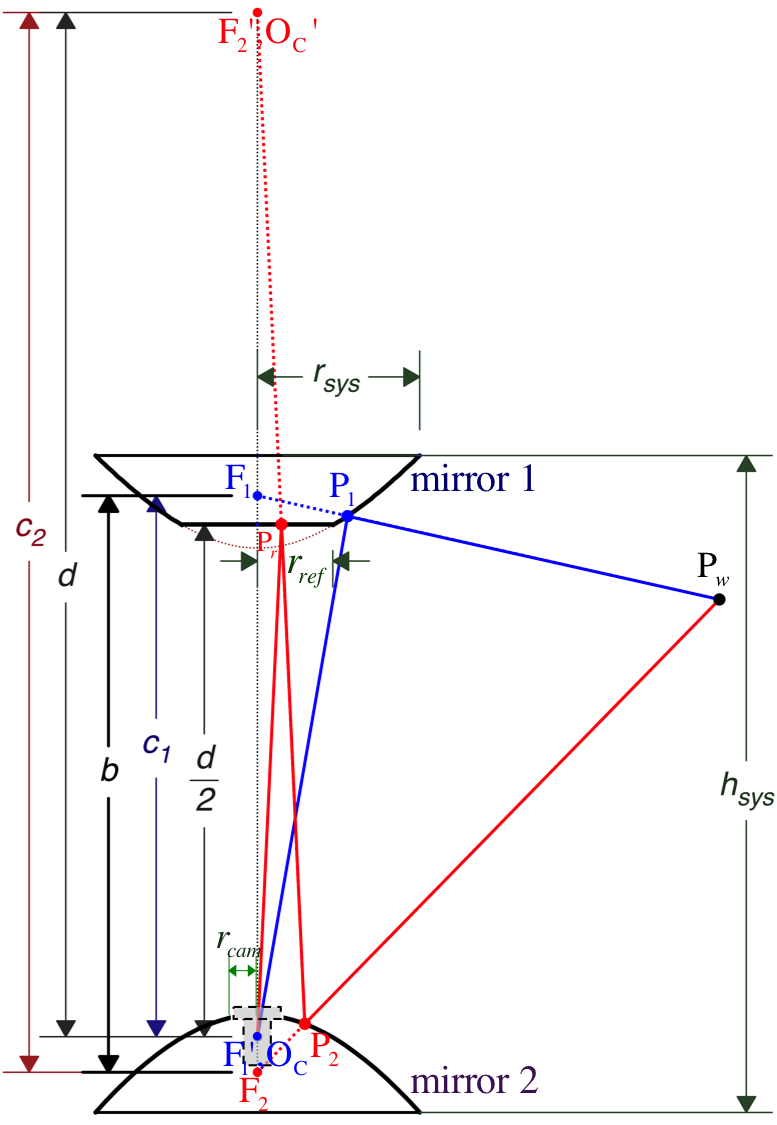}
    \caption[Geometric Model]{2D cross-section of the geometric model and dimension lines for
    observable constraint parameters:
    $c_1, c_2, d, r_{sys}, r_{ref}, r_{cam}$; and relevant by-product parameters:
    baseline $b$ and system height $h_{sys}$.
    In addition, the projection paths corresponding to a world point $\point{P}_w$ are
    also traced in route to the mirrors real foci, $\point{F}_1$ and $\point{F}_2$, and their
    respective reflection points, $\point{P}_1$ and $\point{P}_2$, to the real pinhole camera
    $\point{O_C}$.
    Notice the equivalent projection of $\point{P}_2$ via the virtual camera
    $\point{O_C}'$ achieved by the reflex mirror located at distance $d/2$ from $\point{O_C}$.
    }
    \label{fig:geometric_model}
\end{figure}
In the configuration of \Fref{fig:geometric_model}, mirror 1's real or primary focus is
$\point{F}_1$, which is separated by a distance $c_1$ from its virtual or secondary focus,
$\point{F}_1'$, at the bottom.
Without loss of generality, we make both the camera's pinhole and $\point{F}_1'$
coincide with the origin of the camera's coordinate system, $\point{O_C}$.
This way, the position of the primary
focus, $\point{F}_1$, can be referenced by vector
$\cfs{\vect{f}_1}{}{C}=\vectThreeColInline{0}{0}{c_1}$ in Cartesian coordinates
with respect to the camera frame, $\fr{C}$.
Similarly, the distance between the foci of mirror 2, $\point{F}_2$ and $\point{F}_2'$, is
measured by $c_2$.
Here, we use the planar (reflex) mirror of radius $r_{ref}$ and unit normal
vector 
\begin{align}
    \label{eq:reflex_normal}
    \cfs{\vect{\hat{n}}_{ref}}{}{C}=\vectThreeCol{0}{0}{-1}
\end{align} 
in order to project the real camera's pinhole located at $\point{O_C}$ as a virtual camera
$\point{O_C}'$ coinciding with the virtual focal point $\point{F}_2'$ positioned at
$\cfs{\vect{f}_{2v}}{}{C}=\vectThreeColInline{0}{0}{d}$.
We achieve this by setting $d/2$ as the symmetrical distance from
the reflex mirror to $\point{O_C}$ and from the reflex mirror to $\point{O_C}'$. With
respect to $\fr{C}$, mirror 2's primary focus, $\point{F}_2$, results in position
$\cfs{\vect{f}_2}{}{C}=\vectThreeColInline{0}{0}{d-c_2}$. It yields the following expression for the
reflective plane:
\begin{align}
    \label{eq:reflex_plane}
    \cfs{\trans{\vect{\hat{n}}}_{ref}}{}{C} \cfs{\vect{x}}{}{C}= -d/2
\end{align}

The profile of each hyperboloid is determined by independent parameters $k_1$ and $k_2$,
respectively. Their reflective vertical field of
view (vFOV) are indicated by angles $\alpha_1$ and $\alpha_2$. They play an important role when
designing the total vFOV of the system, $\alpha_{sys}$, formally defined by \fref{eq:vFOV_system}
and illustrated in \Fref{fig:FOVs}. Also importantly, while performing stereo vision,
it is to consider angle $\alpha_{SROI}$, which measures the common (overlapping) vFOV of the
omnistereo system.
The camera's nominal field of view $\alpha_{cam}$ and its opening radius
$r_{cam}$ also determine the physical areas of the mirrors that can be fully imaged.
Theoretically, the mirrors' vertical axis of symmetry (coaxial configuration) produces two image
points that are radially collinear. This property is advantageous for the correspondence search
during stereo sensing (\Fref{sec:3D_sensing}) with a baseline measured as
\begin{equation}
    \label{eq:baseline}
    b = \abs{{c_1} + {c_2} - d}
\end{equation}

Among design parameters, we also include the total height of
the system, $h_{sys}$, and its weight $m_{sys}$, both being formulated in \Fref{sec:size}.

To summarize, the model has 6 primary design parameters given as a vector
\begin{align}
\label{eq:vector_of_primary_design_parameters}
     \vecttheta =\begin{bmatrix}c_1, &c_2, &k_1, &k_2, &d, &r_{sys}\end{bmatrix}
\end{align}
in addition to by-product parameters such as
\begin{align*}
    \begin{bmatrix}
    b, &h_{sys}, &r_{ref}, &r_{cam}, &m_{sys},
    &\alpha_1, &\alpha_2, &\alpha_{sys}, &\alpha_{SROI},
    &\alpha_{cam}
    \end{bmatrix}.
\end{align*}

In \Fref{sec:optimization_and_prototyping}, we perform a numerical
optimization of the parameters in $\vecttheta$ with the goal to maximize the baseline,
$b$, required for life-size navigational stereopsis. At the same time, we
restrict the overall size of the rig (\Fref{sec:size}) without sacrificing sensing performance
characteristics such as vertical field of view, spatial resolution, and depth resolution.
In the upcoming subsections, we first derive the analytical solutions for the
forward projection problem in our coaxial stereo
configuration as a whole. In \Fref{sec:back_projection}, we derive the back-projection
equations for lifting 2D image points into 3D space (up to scale).

\subsection{Single Viewpoint (SVP) Configuration for OmniStereo}
\label{sec:SVP_configuration}
As a central catadioptric system, its projection geometry must obey the existence of the so-called
single effective viewpoint (SVP). While the SVP guarantees that true perspective geometry can always be recovered
from the original image, it limits the selection of mirror profiles to a set of conic sections [2].
Generally, a circular hyperboloid of revolution
(about its axis of symmetry) conforms to the SVP constraint as demonstrated by Baker and Nayar in
\cite{Nayar1997b}.
Since a hyperboloidal mirror has two foci, the effective viewpoint is the primary focus $\point{F}$
inside the physical mirror and the secondary (outer) focus $\point{F'}$ is where the centre
(pinhole) of the perspective camera should be placed for depicting a scene obeying the SVP configuration discussed
in this section.

First of all, a hyperboloid $i$ can be described by the following parametric equation:
\begin{align}
\label{eq:hyperboloid}
    \frac{\left(z_i - z_{0_i}\right)^2}{a_i^2} -\frac{r_i^2}{b_i^2} = 1,
    \;\text{with}\;
    a_i = \frac{c_i}{2}\sqrt{\frac{k_i-2}{k_i}},\; b_i=\frac{c_i}{2}\sqrt{\frac{2}{k_i}}
\end{align}
where the only two parameters are $z_{0_i}=\frac{c_i}{2}$, 
the offset (shift) position of the focus along the Z-axis from the origin $\point{O_C}$, 
and $r_{i}$ is the orthogonal distance to the axis of revolution / symmetry (i.e. the
Z-axis) from a point $\point{P}_i$ on its surface.

In fact, the position of a valid point $\point{P}_i$ is constrained within the mirror's
physical surface of reflection, which is radially limited by $r_{i,min}$ and $r_{i,max}$, such that:
\begin{align}
    \label{eq:r_limits}
    r_i = \sqrt{x_i^2+y_i^2},\; \text{ for } \; r_{i,min} \le r_i \le r_{i,max}, \; \forall i \in \{1, 2\} 
\end{align}
and $ 
r_{1,min} = r_{ref},\; r_{1,max} = r_{sys},\; 
r_{2,min} = r_{cam},\;  r_{2,max} = r_{sys}
$.
Observe that the radius of the system is the upper bound for both mirrors (\Fref{fig:geometric_model}).
In addition, the hyperboloids profiled by
\fref{eq:hyperboloid} must obey the following
conical constraints:
\begin{align}
    \label{eq:hyperbolic_constraints}
    \forall i \in \{1, 2\} \left(c_i > 0 \wedge  k_i > 2 \right)
\end{align}
$k$ is a constant
parameter (unit-less) inversely related to the mirror's curvature or more
precisely, the eccentricity $\varepsilon_c$ of the conic. In fact,
$\varepsilon_c > 1$ for hyperbolas, yet a plane is produced when
$\varepsilon_c \to \infty$ or $k = 2$. 

We devise $M_i$ as the set of all the reflection points
$\point{P}_i$ with coordinates $(x_i, y_i, z_i)$ laying on the surface of the respective mirror $i$
within bounds. Formally,
\begin{align}
\label{eq:set_of_points_on_the_mirror}
    M_i \colonequals \left\{\point{P}_i \in \mathbb{R}^3 \,\middle|\,   \frac{\left(z_i -
    z_{0_i}\right)^2}{a_i^2} -\frac{r_i^2}{b_i^2} = 1 
     \wedge \text{Eq.\eqref{eq:r_limits}} \wedge \text{Eq.\eqref{eq:hyperbolic_constraints}}
    \right\}
\end{align}

In our model, we describe both hyperboloidal mirrors, 1 and 2, with respect to the camera frame
$\fr{C}$, which acts as the common origin of the coordinates system.
Therefore,
\begin{align}
    z_{0_1} &= \frac{c_1}{2} \label{eq:z0_1}
    \\
    z_{0_2} &= d - \frac{c_2}{2} \label{eq:z0_2}
\end{align}

By expanding \fref{eq:hyperboloid} with their respective index terms, it becomes
\begin{align}
 {\left(z_1 - \frac{c_1}{2} \right)^2} - r_1^2\left(\frac{k_1}{2} - 1\right) &=
 \frac{c_1^2}{4}\left(\frac{k_1 - 2}{k_1}\right) \label{eq:hyperboloid1}
 \\
 {\left(z_2 - d + \frac{c_2}{2} \right)^2} -
 r_2^2\left(\frac{k_2}{2} - 1\right) &= \frac{c_2^2}{4}\left(\frac{k_2 - 2}{k_2}\right) \label{eq:hyperboloid2}
\end{align}
Additionally, we define
the function $\func{f}_{z_i}: r \mapsto z_i$ to find the corresponding $z_i$ component
from a given $r$ value as
\begin{align}
    \begin{split}
    \label{eq:func_z} 
    \func{f}_{z_i}(r) &\colonequals  
    \begin{cases}
        z_{0_i} + \gamma_i & \text{if $i=1 \wedge$ Eq.\eqref{eq:r_limits},}
        \\
        z_{0_i} - \gamma_i & \text{if $i=2 \wedge$ Eq.\eqref{eq:r_limits},}
        \\
        None & \text{otherwise.} 
    \end{cases}
    \end{split}
\end{align}
where $\gamma_i = \dfrac{a_i}{b_i} \sqrt{b_i^2 + r_i^2}$.

The inverse relation  $\func{f}_{r_i}: z \mapsto \left\{+r_i, -r_i\right\}$ can be also
implemented as
\begin{align}
    \begin{split}
    \label{eq:func_r}
        \func{f}_{r_i}(z) &\colonequals 
          \begin{cases}
        \pm b_i\Gamma_i  & \text{if $i\in\{1,2\} \wedge$ Eq.\eqref{eq:r_limits},}
        \\
        None & \text{otherwise.} 
    \end{cases}
    \end{split}
\end{align}
where $\Gamma_i = \sqrt{\dfrac{\left(z - z_{0_i}\right)^2}{a_i^2} - 1}$, so 
a valid input $z$ can be associated with both positive and negative solutions $r_i$. 


\subsection{Rig Size}
\label{sec:size}
In the attempt to evaluate the overall system size, we consider the height and
weight variables due to the primary design parameters, $\vecttheta$. 

First, the height of the system, $h_{sys}$ can be estimated from the
functional relationships $\func{f}_{z_1}$ and $\func{f}_{z_2}$ defined in \fref{eq:func_z}, 
which can provide the respective $z-$component values at the out-most point on the mirror's surface.
More specifically, knowing $r_{sys}$, we get
\begin{align}
    \label{eq:height}
    h_{sys} = z_{max} - z_{min} 
\end{align}
where $z_{max} = \func{f}_{z_1}(r_{sys})$ and $z_{min} = \func{f}_{z_2}(r_{sys})$.

The rig's weight can be indicated by the total resulting mass of the main ``tangible" components:
\begin{align}
    \label{eq:mass_sys}
    m_{sys} = m_{cam} + m_{tub} + m_{mir}
\end{align}
where the mass of the camera-lens combination is $m_{cam}$; 
the mass of the support tube can be estimated from its cylindrical volume $V_{tub}$ and material
density $\rho_{tub}$,
and the mass due to the mirrors
\begin{align}
    \label{eq:mass_mirror}
    \begin{split}
        m_{mir} & = V_{mir} \rho_{mir}
        \\ & = \left(V_{1} + V_{ref} + V_{2}\right)  \rho_{mir}
    \end{split}
\end{align}
For computing the volume of the hyperboloidal shell, $V_i$ for mirror $i$, we apply a
``ring method'' of volume integration. By assuming all mirror material has the same wall thickness
$\tau_m$, we acquire $V_i$ by integrating the horizontal cross-sections area along the $\point{Z}$-axis. Each
ring area depends on its outer and inner circumferences that vary according to radius
$r\mid_{z}$ for a given height $z$.
\Fref{eq:func_r} establishes the functional relation
${r_i}^{+}=\func{f}_{r_i}(z)$, from which we only need its positive answer.
We let $A$ be the function that computes the ring area of constant thickness $\tau_m$ for a variable
outer radius $r_i$
\begin{align}
    \label{eq:mirror_area}
    \begin{split}
    A(r_i) &= \pi r_i^2 - \pi\left(r_i-\tau_m\right)^2 
           \\
           &= \pi\tau_m\left(2r_i - \tau_m \right)
    \end{split}
\end{align}
We consider the definite integral evaluated in the $z$ interval bounded by its
height limits, which are correlated with its radial limits \fref{eq:r_limits} and can be obtained
via the $\func{f}_{z_i}$ defined in \fref{eq:func_z}, such that
\begin{align}
    \label{eq:z_limits}
    z_{i,min} = \func{f}_{z_i}\left( r_{i,min} \right)
    \quad \text{ and } \quad
    z_{i,max} = \func{f}_{z_i}\left( r_{i,max} \right)
\end{align}
Then, we proceed to integrate \fref{eq:mirror_area}, so the shell volume for each
hyperboloidal mirror is defined as
\begin{align}
    \label{eq:volume_mirror_hyperboloid}
    V_{i} &= \int ^{z_{i,max} }_{z_{i,min} } A(r_i) \dif z
\end{align}

Finally, since the reflex mirror piece is just a solid cylinder of thickness $\tau_m$, its volume
is simply
\begin{align} 
    \label{eq:volume_reflex_mirror}
    V_{ref}=\tau_m \pi r_{ref}^2
\end{align}

\section{Projective Geometry}
\label{sec:projective_geometry}

\subsection{Analytical Solutions to Projection (Forward)}
\label{sec:forward_projection}

\begin{figure}
    \centering
    \includegraphics[width=0.7\textwidth,keepaspectratio]{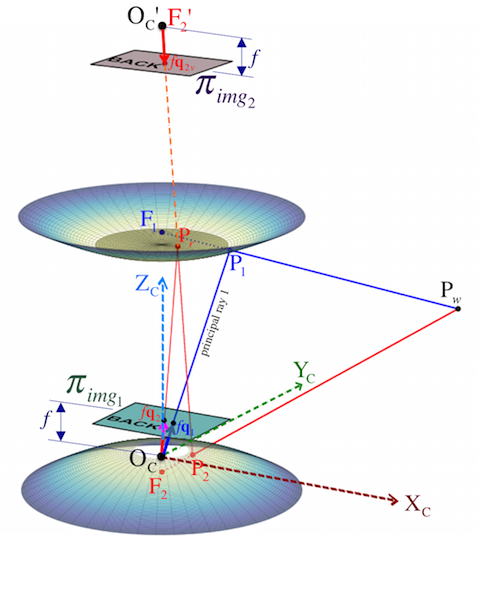}
    \caption[Omnistereo image formation]{Omnistereo image formation. A 3D point $\point{P}_w$ is
    ray-traced as to produce 2D image points (pixels): $\cfs{\pixelPoint{m}_1}{}{I}$ due to
    reflection on mirror 1, and $\cfs{\pixelPoint{m}_2}{}{I}$ due to reflection on mirror 2 via the
    planar (reflex) mirror.
    Although the one-and-only imaging sensor is the camera plane $\pi_{cam}$ 
    , we conveniently imagine two virtual
    projection planes $\pi_{img_1}$ and $\pi_{img_2}$, where the latter presumes to capture points reflecting on mirror 2 
    as if they could pass through the reflex
    mirror toward the virtual camera $\point{O_C}'$. The figure does not show the pixel points
    $\cfs{\pixelPoint{m}_i}{}{I}$ and instead it draws the corresponding vectors
    $f\cfs{\vect{q}_i}{}{C}$ to the points on the projection plane. }
    \label{fig:image_formation}
\end{figure}

Assuming a central catadioptric configuration for the mirrors and camera system (\Fref{sec:SVP_configuration}),
we derive the closed-form solution to the imaging process (forward projection)
for an observable point $\point{P}_w$, positioned in three-dimensional Euclidean space,
$\mathbb{R}^3$, with respect to the reference frame, $\fr{C}$, as vector
$\cfs{\vect{p}_w}{}{C} =\vectThreeColInline{x_w}{y_w}{z_w}$. In addition, we assume all reference
frames such as $\fr{F_1}$ and $\fr{F_2}$ have the same orientation as $\fr{C}$.

For mathematical stability, we must constrain that all projecting world points lie outside
the mirror's volume:
\begin{align}
    \label{eq:external_position_constraint}
    \func{f}_{r_i}(z_{w}) < \rho_{w}, \text{ where $\rho_{w} = \sqrt{x_w^2 + y_w^2}$}
\end{align}
where $\func{f}_{r_i}$ is defined by \fref{eq:func_r} and $\rho_{w}$ measures the horizontal range
to $\point{P}_w$.

$\point{P}_w$ is imaged at pixel position $\cfs{\pixelPoint{m}_1}{}{I}$ after its reflection as
point $\point{P}_1$ on the hyperboloidal surface of mirror 1 (\Fref{fig:image_formation}).  On
the other hand, the second image point's position, $\cfs{\pixelPoint{m}_2}{}{C}$, due to reflection
point $\point{P}_2$ on mirror 2 is rather obtained indirectly after an additional point
$\point{P}_r$ is reflected at $\cfs{\vect{p}_{ref}}{}{C}$ on the reflex mirror represented via
\fref{eq:p_reflex}.

First, for $\point{P}_w$'s reflection point via mirror 1 at
position vector $\cfs{\vect{p}_1}{}{C}$, 
we use $\lambda_1$ as the parametrization term for the line equation 
passing through $\point{F}_1$ toward $\point{P}_w$ with direction   
$\cfs{\vect{d}_1}{}{F_1}=\cfs{\vect{p}_w}{}{C} - \cfs{\vect{f}_1}{}{C}$. The position of any
point $\point{P}_1$ on this line is given by:
\begin{equation}
    \label{eq:line_f1_pw}
    \cfs{\vect{p}_1}{}{C}
    =\cfs{\vect{f}_1}{}{C} + \lambda_1 \cfs{\vect{d}_1}{}{F_1}
\end{equation}

Substituting \fref{eq:line_f1_pw} into \fref{eq:hyperboloid1}, we obtain:
\begin{align*}
    \left( \lambda_1 (z_w - c_1) + \frac{c_1}{2} \right)^2 
    - \left( \lambda_1^2 x_w^2 + \lambda_1^2 y_w^2\right)
    \left(\frac{{{k_1}}}{2} - 1\right) &
    \\ 
    - \frac{{c_1^2}}{4}\left(\frac{{{k_1} - 2}}{{{k_1}}}\right) &= 0
\end{align*}
in order to solve for $\lambda_1$, which turns out to be
\begin{equation}
    \label{eq:lambda_1}
    \lambda_1=\frac{c_1}{\norm{\cfs{\vect{d}_1}{}{F_1}}\sqrt{k_1\cdot(k_1 - 2)} 
     - k_1 \left(z_w - c_1\right)}
\end{equation} 
where $\norm{\cfs{\vect{d}_1}{}{F_1}}
=\sqrt{x_w^2+y_w^2+(z_w-c_1)^2}$ is the Euclidean norm between $\point{P}_w$ and mirror
1's focus, $\point{F}_1$.

In practice, we represent the reflection point's position $\cfs{\vect{p}_{1}}{}{C}$
as a matrix-vector multiplication between the $3\times4$ transformation matrix 
$\matx{K}_{1} = \vectTwoCol{\lambda_1 \matx{I}_{(3)}}{\left(1 - \lambda_1
\right)\cfs{\vect{f}_1}{}{C}}$ and the point's position vector
$\cfs{\vect{p}_{w,h}}{}{C}=\vectFourColInline{x_w}{y_w}{z_w}{1}$ in homogeneous coordinates:
\begin{equation}
    \label{eq:p1}
    \cfs{\vect{p}_{1}}{}{C}
    = \matx{K}_{1} \cfs{\vect{p}_{w,h}}{}{C}
\end{equation}
Note that $\cfs{\vect{p}_1}{}{C}$'s elevation angle, $\theta_1$, must be bounded as 
\begin{align}
    \label{eq:theta1_limits}
    \theta_{1,min} \le \theta_1 \le \theta_{1,max}
\end{align}
where $\theta_{1,min}$ and
$\theta_{1,max}$ are the angular elevation limits 
for the real reflective area of the hyperboloid. 


Finally, the reflection point $\point{P}_1$ with position
$\cfs{\vect{p}_{1}}{}{C}$ can
now be perspectively projected as a pixel point located at $\cfs{\pixelPoint{m}_1}{}{I}=
\vectTwoColInline{u_1}{v_1}$ on the image.
In fact, the entire imaging process of $\point{P}_w$ via mirror 1 can be expressed in homogeneous
coordinates as:
\begin{equation}
    \label{eq:m1}
    \cfs{\pixelPoint{m}_{1,h}}{}{I} 
    =
    \zeta_1 
    \matx{K}_{c}
    \matx{K}_{1}
    \cfs{\vect{p}_{w,h}}{}{C}
\end{equation}
where the scalar
$\zeta_1 = 1/z_{1} = 1/ \left(c_{1} + \lambda_{1} \left( z_w -
c_1\right) \right)$
is the perspective normalizer that maps the {\em principal ray}
passing through $\vect{p}_{1}$ onto a point
$\cfs{\vect{q}_{1}}{}{C}=\vectThreeColInline{x_{q_1}}{y_{q_1}}{1}$ on the normalized projection
plane $\hat{\pi}_{img_1}$. The
traditional $3\times3$ intrinsic matrix of the camera's pinhole model is
\begin{align}
    \label{eq:camera_intrinsic_matrix}
    \matx{K}_{c} =
    \begin{bmatrix}
    f_u & s   & u_c \\
    0   & f_v & v_c \\
    0   & 0   & 1 
    \end{bmatrix}
\end{align}
in which
$f_u=f/h_x$ and $f_v=f/h_y$ are based on the focal length $f$
and the pixel dimension $(h_x, h_y)$, $s$ is the skew parameter, and
$\cfs{\pixelPoint{m}_c}{}{I}=\vectTwoColInline{u_c}{v_c}$ is the optical center position on the
image $\fr{I}$ 
\Fref{fig:image_formation} illustrates the projection point $f\cfs{\vect{q}_1}{}{C}$ on the
respective image plane $\pi_{img_1}$.

Similarly, we provide the analytical solution for the forward projection of $\point{P}_w$ via mirror
2 by first considering the position of reflection point $\point{P}_2$:
\begin{equation}
    \label{eq:p2}
    \cfs{\vect{p}_{2}}{}{C}
    = \matx{K}_{2} \cfs{\vect{p}_{w,h}}{}{C}
\end{equation} 
where $\matx{K}_{2} = \vectTwoCol{\lambda_2 \matx{I}_{(3)}}{\left(1 - \lambda_2
\right)\cfs{\vect{f}_2}{}{C}}$ is similar to the transformation matrix $\matx{K}_1$, but obviously
it now uses $\cfs{\vect{f}_2}{}{C}$ and
\begin{equation}
    \label{eq:lambda_2}
    \lambda_2=\frac{c_2}{\norm{\cfs{\vect{d}_2}{}{F_2}}\sqrt{k_2\cdot(k_2 - 2)} 
    + k_2 \left(z_w - (d-c_2) \right)} 
\end{equation} 
with direction vector's norm 
\begin{align}
\label{eq:d2_mag}
\norm{\cfs{\vect{d}_2}{}{F_2}} = \norm{\cfs{\vect{p}_w}{}{C}-\cfs{\vect{f}_2}{}{C}}
=\sqrt{x_w^2+y_w^2+\left(z_w-(d-c_2)\right)^2}
\end{align}
For completeness, note that the physical projection via mirror 2 is incident to the reflex mirror at
\begin{align}
    \label{eq:p_reflex}
    \cfs{\vect{p}_{ref}}{}{C} = \cfs{\vect{f}_{2v}}{}{C} +
    \lambda_{ref}\left(\cfs{\vect{p}_2}{}{C}-\cfs{\vect{f}_{2v}}{}{C}\right)
\end{align}
where $\lambda_{ref} = \frac{d}{2(d-z_2)}$ according to
\fref{eq:reflex_plane} in the theoretical model.
Ultimately, ignoring any astigmatism and
chromatic aberrations introduced by the reflex mirror, and because the
same (and only) real camera with $\matx{K}_{c}$ is used for imaging, we
obtain the projected pixel position $\cfs{\pixelPoint{m}_{2,h}}{}{I}=
\vectThreeColInline{u_2}{v_2}{1}$ in homogeneous coordinates:
\begin{equation}
    \label{eq:m2}
    \cfs{\pixelPoint{m}_{2,h}}{}{I}
    =
    \zeta_2 
    \matx{K}_{c}
    \matx{M}_{ref}
    \matx{K}_{2}
    \cfs{\vect{p}_{w,h}}{}{C}
\end{equation}
where $\zeta_2 = 1/\left(d-z_2\right) 
$ 
is the perspective
normalizer to find $\cfs{\vect{q}_2}{}{C}$ on the normalized
projection plane, $\hat{\pi}_{img_2}$.

Due to planar mirroring via the reflex mirror,
$\cfs{\matx{M}_{ref}}{C}{C'}$ is used to change the coordinates of
$\point{P}_2$ from $\fr{C}$ onto the virtual camera frame,
$\fr{C'}$, located at $\cfs{\vect{f}_{2v}}{}{C}$.
Hence, 
\begin{align}
    \label{eq:frame_change_C_to_Cvirt}
    \cfs{\matx{M}_{ref}}{C}{C'} = 
    \begin{bmatrix}
        \matx{I}_{(3)} + 2 \matx{D}_{\vect{\hat{n}}_{ref}}, && \cfs{\vect{f}_{2v}}{}{C}
    \end{bmatrix}
\end{align}
where the $3\times1$ unit normal vector of the reflex mirror plane,
$\cfs{\vect{\hat{n}}_{ref}}{}{C}$ given in \eqref{eq:reflex_normal}, is mapped into its
corresponding $3\times3$ diagonal matrix $\matx{D}_{\vect{\hat{n}}_{ref}}$, via the relationship:
\begin{align}
    \label{eq:matrix_form_of_normal_vector}
    \matx{D}_{\vect{\hat{n}}_{ref}} \gets 
    \matx{I}_{(3)} \func{diag} \! \left( \cfs{\vect{\hat{n}}_{ref}}{}{C}
    \right)
\end{align}

It is convenient to define the forward projection functions
$\funcvv{f}_{\varphi_1}(\cfs{\vect{p}}{}{C})$ and $\funcvv{f}_{\varphi_2}(\cfs{\vect{p}}{}{C})$ for
a 3D point $\point{P}$ whose position vector is known with respect to $\fr{C}$ and which is situated
within the vertical field of view $\alpha_{i}$ of mirror $i$ (for $i\in\{1,2\}$) indicated in
\Fref{fig:FOVs}.
The respective function $\funcvv{f}_{\varphi_i}(\cfs{\vect{p}}{}{C})$ maps to image point $\cfs{\pixelPoint{m}_i}{}{I}$ on
frame $\fr{I}$, such that $\funcvv{f}_{\varphi_i}\colon \mathbb{R}^3 \mapsto \mathbb{R}^2$ and
implemented as:
\begin{align}
    \label{eq:forward_projection_function}
    \funcvv{f}_{\varphi_i}(\cfs{\vect{p}}{}{C}) &\colonequals
    \begin{cases}
        \cfs{\vect{p}}{}{C} \xmapsto{\text{Eq. \eqref{eq:m1}}} \cfs{\pixelPoint{m}_1}{}{I}
        &\! \text{if $i=1 \wedge$ Eqs.
        \eqref{eq:image_radial_bounds}\eqref{eq:external_position_constraint},}
        \\
        \cfs{\vect{p}}{}{C} \xmapsto{\text{Eq. \eqref{eq:m2}}} \cfs{\pixelPoint{m}_2}{}{I}
        &\! \text{if $i=2 \wedge$ Eqs.
        \eqref{eq:image_radial_bounds}\eqref{eq:external_position_constraint},}
        \\
        None &\! \text{otherwise.}
    \end{cases}
\end{align}
In fact, $\cfs{\pixelPoint{m}_i}{}{I}$ is considered valid if it is located within the imaged
radial bounds, such that:
\begin{align}
    \label{eq:image_radial_bounds}
    ~\cfs{\mynorm*{\cfs{\pixelPoint{m_{r_{i,min}}}}{}{I}}}{}{I_{C_i}}
    ~\le~
    \cfs{\mynorm*{\cfs{\pixelPoint{m_i}}{}{I}}}{}{I_{C_i}}
    ~\le~
    \cfs{\mynorm*{\cfs{\pixelPoint{m_{r_{i,max}}}}{}{I}}}{}{I_{C_i}}
\end{align}
where the frame of reference $\fr{I_{C_i}}$ implies that its origin is the image center
$\cfs{\pixelPoint{m}_c}{}{I_i} = \vectTwoColInline{u_{c_i}}{v_{c_i}}$ of the $\fr{I_i}$ masked
imaged.
Therefore, the magnitude (norm) of any position $\cfs{\pixelPoint{m}}{}{I_{C_i}}$ in pixel space
$\fr{I_{C_i}}$ can be measured as
\begin{align}
    \label{eq:radial_pixel_magnitude}
    {\cfs{\mynorm{\cfs{\pixelPoint{m}}{}{I_i}}}{}{I_{C_i}}} \colonequals
    {\mynorm{\cfs{\pixelPoint{m}}{}{I_i}-\cfs{\pixelPoint{m}_c}{}{I_i}} } =
    {\sqrt{(u-u_c)^2+(v-v_c)^2}}
\end{align}
In particular, $\cfs{\mynorm*{\cfs{\pixelPoint{m_{r_{i,lim}}}}{}{I}}}{}{I_{C_i}}$
is the image radius obtained from the projection $\cfs{\pixelPoint{m_{r_{i,lim}}}}{}{I} \gets
\funcvv{f}_{\varphi_i}(\cfs{\vect{p}_{i,lim}}{}{C})$ corresponding to a particular point
coincident with the line of sight of the radial limit $r_{i,lim}$ -- it being either
$r_{sys}$, $r_{ref}$, or $r_{cam}$ as indicated by \fref{eq:r_limits}.

\subsection{Analytical Solutions to Back Projection}
\label{sec:back_projection}

The back projection procedure establishes the relationship between the 2D position of a pixel
point $\cfs{\pixelPoint{m}}{}{I}=\vectTwoColInline{u}{v}$ on the image $\fr{I}$ and its corresponding
3D projective direction vector $\vect{v}$ toward the observed point $\point{P}_w$ in the world.

Initially, the pixel point $\cfs{\pixelPoint{m}_1}{}{I}$ (imaged via mirror 1) is mapped as
$\point{Q}_1$ onto the normalized projection plane $\hat{\pi}_{img_1}$ with coordinates 
$\cfs{\vect{q}_1}{}{C} = \vectThreeColInline{x_{q_1}}{y_{q_1}}{1}$ by applying the inverse
transformation of the camera intrinsic matrix \fref{eq:camera_intrinsic_matrix} as follows:
\begin{equation}
    \label{eq:q1_bp}
    \cfs{\vect{q}_1}{}{C}
    =
    \cfs{\matx{K}^{-1}_{c}}{I}{C} \cfs{\vect{m}_{1,h}}{}{I}
    =
    \begin{bmatrix}
    \frac{1}{f_u} && -\frac{s}{f_u f_v}   && \frac{s v_c - f_v u_c}{f_u f_v} \\
    0             && \frac{1}{f_v}        && - \frac{v_c}{f_v} \\
    0             && 0                    && 1 
    \end{bmatrix}
    \begin{bmatrix}
    u_1 \\
    v_1 \\
    1    
    \end{bmatrix}
\end{equation}

For simplicity, we assume no distortion parameters exist, so we can proceed with
the lifting step along the principal ray that passes through three points: the camera's pinhole 
$\point{O_C}$, point $\point{Q}_1$ on the projection plane,
and the reflection point $\point{P}_1$ (\Fref{fig:image_formation}). The vector form of this line
equation can be written as:
\begin{align}
    \label{eq:ray_p1_bp}
    \cfs{\vect{p}_{1}}{}{C}
    & =\cfs{\vect{o}_c}{}{C} + t_1\left(\cfs{\vect{q}_1}{}{C} - \cfs{\vect{o}_c}{}{C} \right) 
    = t_1 \cfs{\vect{q}_1}{}{C}
\end{align}

By substituting
\fref{eq:ray_p1_bp} into \eqref{eq:hyperboloid1}, we solve for the parameter $t_1$, to get
\begin{equation}
    \label{eq:t_1}
    t_1=\frac{c_1}{k_1 - \norm{\cfs{\vect{q}_1}{}{C}}\sqrt{k_1\cdot(k_1 - 2)}}
\end{equation}
where $\norm{\cfs{\vect{q}_1}{}{C}}
=\sqrt{x_{q_1}^2+y_{q_1}^2+1}$ is the distance between $\point{Q}_1$ and
$\point{O_C}$.



Given $\cfs{\vect{v}_1}{}{F_1}$ as the direction
vector leaving focal point $\point{F}_1$ toward the world point $\cfs{\point{P}_w}{}{C}$. Through 
frame transformation $ \cfs{T_1}{C}{F_1}\cfs{\vect{p}_{1,h}}{}{C}$, we get
\begin{align}
    \label{eq:bp_ray1_wrt_F1}
    \cfs{\vect{v}_1}{}{F_1} =
    \cfs{T_1}{C}{F_1}\cfs{\vect{p}_{1,h}}{}{C}
    ~\text{, where} ~
    \cfs{T_1}{C}{F_1}_{(3\times4)}=\begin{bmatrix}
                                    \matx{I}_{(3)}, && - \cfs{\vect{f}_1}{}{C}
                                \end{bmatrix}
\end{align}
for $\cfs{\vect{p}_{1,h}}{}{C}$ as the
homogeneous form of \eqref{eq:ray_p1_bp}.
In fact, $\cfs{\vect{v}_1}{}{F_1}$ provides the back-projected angles (elevation $\theta_1$,
azimuth $\psi_1$) from focus $\point{F}_1$ toward $\cfs{\point{P}_w}{}{C}$:
\begin{align}
        \cfs{\theta_1}{}{F_1} &= 
        \asin{\frac{z_{\vect{v}_1}}{\norm{\cfs{\vect{v}_1}{}{F_1}}}}
        =
        \asin{\frac{z_1 - c_1}{\norm{\cfs{\vect{v}_1}{}{F_1}}} }
        \label{eq:theta1}
    \\
        \cfs{\psi_1}{}{F_1} &= \atan{\frac{y_{\vect{v}_1}}{x_{\vect{v}_1}}}
        = \atan{\frac{y_1}{x_1}}
        \label{eq:azimuth1}
\end{align}
where $\mynorm*{\cfs{\vect{v}_1}{}{F_1}}$ is the norm of the back-projection vector up to the mirror
surface.

Using the same approach, we lift a pixel point $\cfs{\pixelPoint{m}_2}{}{I}$ imaged via mirror 2.
Because the virtual camera $\point{O_C}'$ located at
$\cfs{\vect{f}_2}{}{C}=\vectThreeColInline{0}{0}{d-c_2}$ uses the same intrinsic matrix $\matx{K}_{c}$, 
we can safely back-project pixel $\cfs{\pixelPoint{m}_2}{}{I}$ to
$\point{Q}_{2v}$ on the normalized projection plane $\hat{\pi}_{img_2}$ as follows:
\begin{align}
    \label{eq:q_2v_wrt_virtual_camera_frame}
    \cfs{\vect{q}_{2v}}{}{C'} = \cfs{\vect{q}_{2}}{}{C} =
    \matx{K}^{-1}_{c}\cfs{\pixelPoint{m}_{2,h}}{}{I}
\end{align}
where the inverse transformation of the camera intrinsic matrix
$\matx{K}^{-1}_{c}$ is given by \fref{eq:camera_intrinsic_matrix}.
Since the reflection matrix $\matx{M}_{ref}$ defined in
\fref{eq:frame_change_C_to_Cvirt} is bidirectional due to the symmetric position of the reflex
mirror about $\fr{C}$ and $\fr{C'}$, we can find the desired position of $\cfs{\vect{q}_{2v}}{}{C}$
with respect to $\fr{C}$:
\begin{equation}
    \label{eq:q2v_bp_wrt_C}
    \cfs{\vect{q}_{2v}}{}{C} = \cfs{\matx{M}_{ref}}{C'}{C} \cfs{\vect{q}_{2v,h}}{}{C'} 
\end{equation}
which is equivalent to 
$\cfs{\vect{q}_{2v}}{}{C} = \vectThreeColInline{x_{q_{2v}}}{y_{q_{2v}}}{d-1}$.

In \Fref{fig:image_formation}, we can see the principal ray that passes through the virtual camera's
pinhole $\point{O_C}'$ and the reflection point $\point{P}_2$, so this line equation can be written
as:
\begin{align}
    \label{eq:ray_p2_bp}
    \cfs{\vect{p}_{2}}{}{C}
    & = \cfs{\vect{f}_{2v}}{}{C} + t_2\left(\cfs{\vect{q}_{2v}}{}{C} - \cfs{\vect{f}_{2v}}{}{C}
    \right)
\end{align}

Solving for $t_2$ from 
\fref{eq:ray_p2_bp} and \eqref{eq:hyperboloid2}, we get
\begin{equation} 
    \label{eq:t_2}
    t_2=\frac{c_2}{k_2-\norm{\cfs{\vect{q}_2}{}{C}}\sqrt{k_2\cdot(k_2 - 2)}}
\end{equation}
where $\norm{\cfs{\vect{q}_2}{}{C}}
=\sqrt{x_{q_2}^2+y_{q_2}^2+1}$ is the distance between the
normalized projection point $\point{Q}_2$ and the camera $\point{O_C}$ while considering
\eqref{eq:q2v_bp_wrt_C}.
Beware that the newly found location of $\point{P}_2$ is given with respect to
the real camera frame, $\fr{C}$.


Again, we obtain the back-projection ray
\begin{align}
    \label{eq:bp_ray2_wrt_F2}
    \cfs{\vect{v}_2}{}{F_2} = \cfs{T_2}{C}{F_2}\cfs{\vect{p}_{2,h}}{}{C}
    ~\text{, where}~
    \cfs{T_2}{C}{F_2}_{(3\times4)}=\begin{bmatrix}
                        \matx{I}_{(3)}, && -\vect{f}_2
    \end{bmatrix}
\end{align}
in order to indicate the direction leaving from the primary focus $\point{F}_2$ toward $\point{P}_w$
through $\point{P}_2$.
Here, the corresponding elevation and azimuth angles are respectively given by
\begin{align}
    \cfs{\theta_2}{}{F_2} &= 
    \asin{\frac{z_{\vect{v}_2}}{\norm{\cfs{\vect{v}_2}{}{F_2}}}}
    =
    \asin{\frac{d - t_2}{\norm{\cfs{\vect{v}_2}{}{F_2}}}}
    \label{eq:theta2}
\\
    \cfs{\psi_2}{}{F_2} &= \atan{\frac{y_{\vect{v}_2}}{x_{\vect{v}_2}}}
    = \atan{\frac{y_2}{x_2}}
    \label{eq:azimuth2}
\end{align}
where $\mynorm*{\cfs{\vect{v}_2}{}{F_2}}=\sqrt{x_2^2 + y_2^2 +
\left(c_2-t_2\right)^2}$ is the magnitude of the direction vector from its reflection point
$\point{P}_2$.
 
Like done for the (forward) projection, it is convenient to define the back-projection
functions $\funcvv{f}_{\beta_1}$ and
$\funcvv{f}_{\beta_2}$ for lifting a 2D pixel point
$\cfs{\pixelPoint{m}}{}{I}$ within valid radial bounds \fref{eq:image_radial_bounds} to their
angular components $\cfs{\left(\theta_i, \psi_i\right)}{}{F_i}$ with respect to their respective
foci frame $\fr{F_i}$ (oriented like $\fr{C}$) as indicated by equations
\eqref{eq:theta1},\eqref{eq:azimuth1} and \eqref{eq:theta2},\eqref{eq:azimuth2}, such that $\funcvv{f}_{\beta_i}\colon \mathbb{R}^2 \mapsto \mathbb{R}^2$,
which is implemented as follows:
\begin{align}
    \label{eq:back_projection_function}
    \funcvv{f}_{\beta_i}(\cfs{\pixelPoint{m}}{}{I}) &\colonequals
    \begin{cases}
        \left(
        \cfs{\pixelPoint{m}}{}{I} \xmapsto{\text{\eqref{eq:theta1}}}\cfs{\theta_1}{}{F_1},
        \cfs{\pixelPoint{m}}{}{I} \xmapsto{\text{\eqref{eq:azimuth1}}} \cfs{\psi_1}{}{F_1}
        \right)
        & \text{if $i=1$,}
        \\
        \left(
        \cfs{\pixelPoint{m}}{}{I} \xmapsto{\text{\eqref{eq:theta2}}} \cfs{\theta_2}{}{F_2},
        \cfs{\pixelPoint{m}}{}{I} \xmapsto{\text{\eqref{eq:azimuth2}}}\cfs{\psi_2}{}{F_2}
        \right)
        & \text{if $i=2$,}
        \\
        None & \neg \text{\eqref{eq:image_radial_bounds}.}
    \end{cases}
\end{align}

\subsection{Field-of-View}
\label{sec:FOV}
\begin{figure}
    \centering
    \includegraphics[width=0.8\textwidth,keepaspectratio]{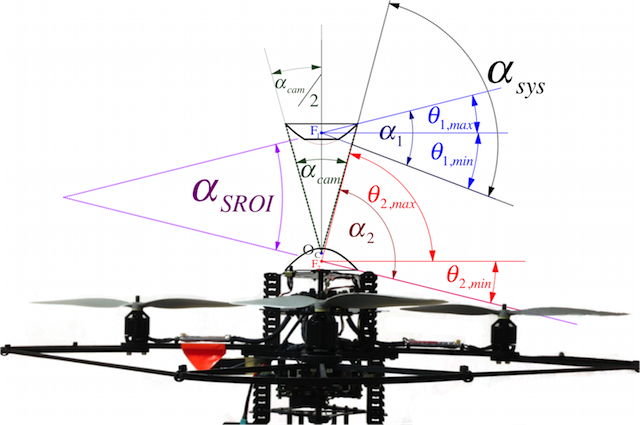}
    \caption[Vertical Field of View (vFOV) angles]{Vertical Field of View (vFOV) angles: $\alpha_1$
    and $\alpha_2$ are the individual vFOV angles of the mirrors formed by their
    respective limits $\theta_{1/2,min/max}$; $\alpha_{sys}$ is the overall vFOV
    angle of the system; and $\alpha_{SROI}$ indicates the angle of the
    overlapping region conceived from vFOVs $\alpha_1$ and $\alpha_2$.}
    \label{fig:FOVs}
\end{figure}
The horizontal FOV is clearly \ang{360} for both mirrors. In other words, azimuths $\psi$ can be
measured in the interval $\left[0, 2\pi \right )$ \si{\radian}.

As discussed previously, there exists a positive correlation between the vertical field of view
(vFOV) angle $\alpha_i$ of mirror $i$ and its profile parameter $k_i$, such that $\alpha_i \to
\ang{180}$ as $k_i \to \infty$ (see \Fref{fig:effect_of_k_on_system_radius}).
As demonstrated in \Fref{fig:FOVs}, $\alpha_i$ is physically bounded by its corresponding elevation
angles:
$\theta_{i,max}$, $\theta_{i,min}$. Both vFOV angles, $\alpha_1$ and $\alpha_2$, are computed
from their elevation limits as follows:
\begin{subequations}
    \label{eq:alpha12}
    \begin{align}
        \alpha_1 &=  \theta_{1,max}  - \theta_{1,min}         \label{eq:alpha1}
        \\
        \alpha_2 &=  \theta_{2,max}  - \theta_{2,min}         \label{eq:alpha2}
    \end{align}
\end{subequations}
In fact, the overall vFOV of the system is also given from these elevation
limits:
\begin{equation}
    \label{eq:vFOV_system}
    \alpha_{sys} = \max \left(\theta_{1,max},
    \theta_{2,max}\right) - \min \left(\theta_{1,min},
    \theta_{2,min}\right)
\end{equation}

\begin{figure}
    \centering \includegraphics[width=0.8\textwidth,keepaspectratio]{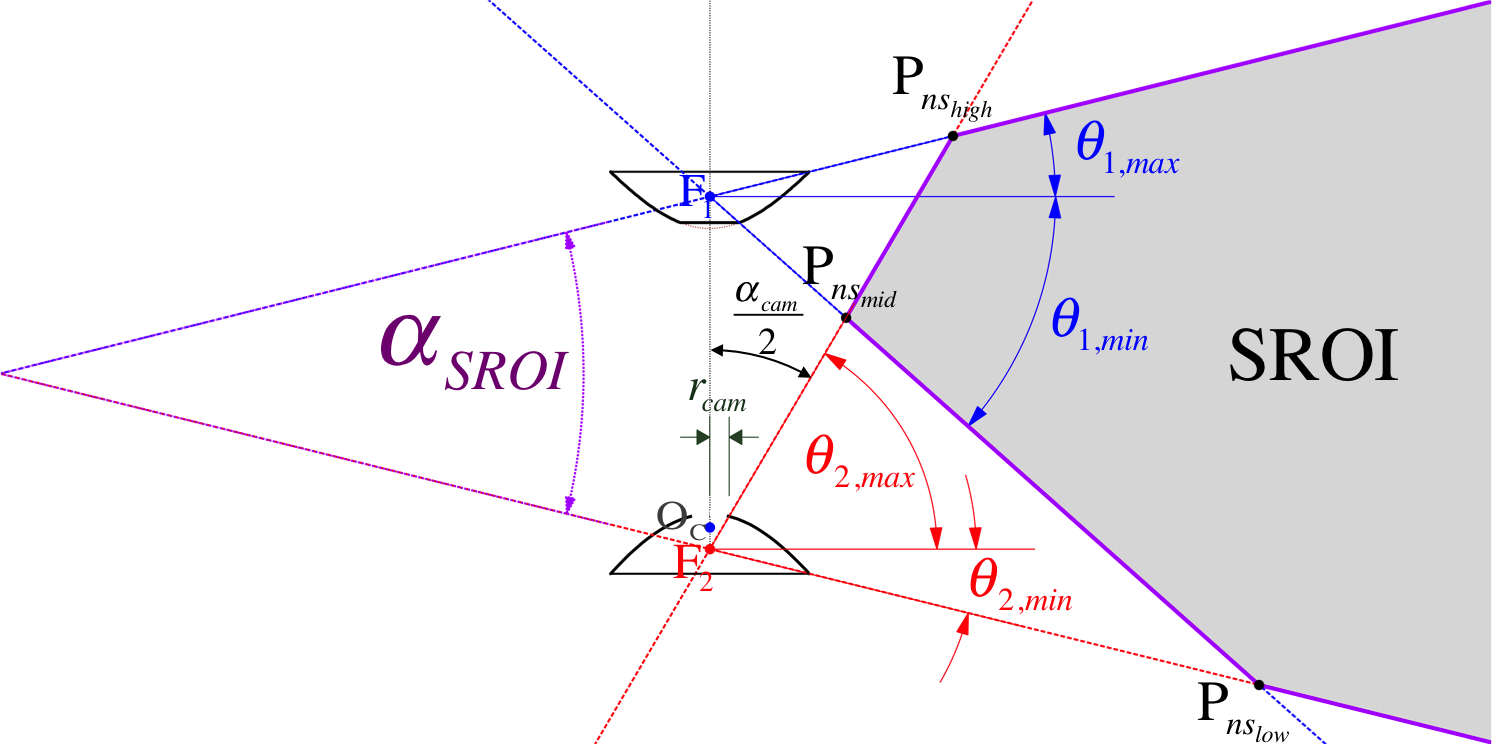}
    \caption[Common vFOV and Stereo ROI]{A cross section of the Stereo ROI (shaded area) formed by
    the intersection of view rays for the limiting elevations $\theta_{1/2,min/max}$. The nearest
    stereo ($ns$) points are labeled $\point{P}_{ns_{high}}$, $\point{P}_{ns_{mid}}$ and
    $\point{P}_{ns_{low}}$ since they are the vertices of the hull that near-bounds the set of
    usable points for depth computation from triangulation (\Fref{sec:triangulation}). 
    See the \Fref{tab:vertices_stereo_ROI} for an example.
    }
    \label{fig:common_FOV_ROI}
\end{figure}

\Fref{fig:common_FOV_ROI} highlights the the so-called common vFOV angle, $\alpha_{SROI}$, for the
overlapping region of interest (Stereo ROI) where the same point can be seen from
both mirrors so stereo computation can be performed (\Fref{sec:3D_sensing}).
In our model, $\alpha_{SROI}$ can be decided from the value of the three prevailing elevation angles
($\theta_{1,max}$, $\theta_{1,min}$, and $\theta_{2,min}$), such that:
\begin{equation}
    \label{eq:vFOV_common}
    \alpha_{SROI} = \theta_{SROI,max} - \theta_{SROI,min}
\end{equation}
where generally, 
\begin{subequations}
    \label{eq:common_FOV_limit_angles}
    \begin{align}
        \theta_{SROI,min} &= \max(\theta_{1,min},\theta_{2,min})
        \label{eq:common_FOV_min_angle}
        \\
        \theta_{SROI,max} &= \min(\theta_{1,max},\theta_{2,max})
        \label{eq:common_FOV_max_angle}
    \end{align} 
\end{subequations}

The shaded area in \Fref{fig:common_FOV_ROI} illustrates the Stereo ROI 
that is far-bounded by the set of triangulated points found at the maximum range due to minimum disparity
$\Delta m_{12}=1\,\mathrm{px}$ in the discrete case (refer to
\Fref{fig:horizontal_range_variation_due_to_pixel_disparity}), such that
\begin{align}
\label{eq:max_horizontal_range_points_set}
    \begin{split}
    P_{fs} = 
    \left\{
    \right.
    \point{P}_w \gets \funcvv{f}_{\Delta}
    ((\theta_1,\psi_1), (\theta_2,\psi_2) )
    \mid&
    (\theta_1, \psi_{1}) \gets \funcvv{f}_{\beta_1}(\pixelPoint{m_1}) 
    \\
    \wedge&
    (\theta_2, \psi_{2}) \gets \funcvv{f}_{\beta_2}(\pixelPoint{m_2}) 
    \\ 
    \wedge&
    \left.
    \Delta m_{12}= 1,\mathrm{px}
    \right\}
    \end{split}
\end{align}
where functions $\funcvv{f}_{\beta_i}$ and $\funcvv{f}_{\Delta}$,
are provided in \eqref{eq:back_projection_function} and \eqref{eq:function_triangulation}. 

The Stereo ROI is near-bounded (to the $\point{Z}$-axis of radial symmetry) by its
vertices $\point{P}_{ns_{high}}$, $\point{P}_{ns_{mid}}$ and $\point{P}_{ns_{low}}$, which result 
from the following ray-intersection cases:
\begin{enumerate}[a)]
    \item $\point{P}_{ns_{high}} \gets
    \funcvv{f}_{\Delta}((\theta_{1,max},\psi_1),(\theta_{2,max},\psi_2))$
    \item $\point{P}_{ns_{mid}} \gets \funcvv{f}_{\Delta}((\theta_{1,min},\psi_1),(\theta_{2,max},
    \psi_2))$
    \item $\point{P}_{ns_{low}} \gets \funcvv{f}_{\Delta}((\theta_{1,min},\psi_1),(\theta_{2,min},
    \psi_2))$
\end{enumerate}
where (again) the intersection function $\funcvv{f}_{\Delta}$ is implemented for direction
rays (or angles) as defined in the Triangulation \Fref{sec:triangulation}.


By assuming a radial symmetry on the camera's field of view $\alpha_{cam}$, it should allow for a
complete view of the mirror surface at its outmost diameter of $2r_{1,max} = 2r_{sys}$ according to
\fref{eq:r_limits}. Substantially, as depicted in \Fref{fig:common_FOV_ROI}, $\alpha_{cam}$ is
upper-bounded by the camera hole radius $r_{cam}$ selected according to \fref{eq:r_cam}. 
The following inequality constraint emerges
\begin{align}
\label{eq:camera_FOV_nonequality_constraint}
   2\atan{\frac{r_{sys}}{ \func{f}_{z_1}(r_{sys})}} 
   \le \alpha_{cam} \le 
   2 \atan{\frac{r_{cam}}{\func{f}_{z_2}(r_{cam})}}
\end{align}
where the respective function $\func{f}_{z_i}$ is defined in \eqref{eq:func_z}.

\begin{figure}
    \centering
    \includegraphics[width=0.8\textwidth,keepaspectratio]{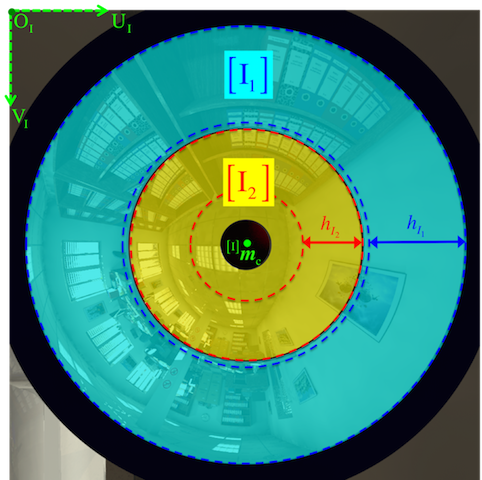}
    \caption[Masked images.]{The omnidirectional image $\fr{I}$ shown in
    \Fref{fig:sim_rig_camera_view} is now annotated for the separate regions of interest in
    $\fr{I_1}$ and $\fr{I_2}$ from the corresponding projections onto their image planes
    $\pi_{img_i}$.
    In addition, we indicate the corresponding radial heights $h_{I_1}$ and $h_{I_2}$ for the Stereo
    ROI that are used determine the imaging ratio $\chi_{\point{I}_{1:2}}=\frac{h_{I_1}}{h_{I_2}}$.
    }
    \label{fig:imaging_ratio}
\end{figure} 
Our specific viewing requirements when mounting the
omnidirectional sensor along the central axis of the quadrotor ensure that objects
located at \SI{15}{\centi\metre} under the rig's base and at 1 meter away
(from the central axis) can be viewed. 
Thus, angles $\theta_{1,min}$ and $\theta_{2,min}$ should only be large enough
as to avoid occlusions from the MAV's propellers (\Fref{fig:FOVs}) and
to produce inner and outer ring images at useful ratio (\Fref{fig:imaging_ratio}).

\subsection{Spatial Resolution}
\label{sec:spatial_resolution}
\begin{figure}
    \centering
    \includegraphics[width=0.8\columnwidth,keepaspectratio]{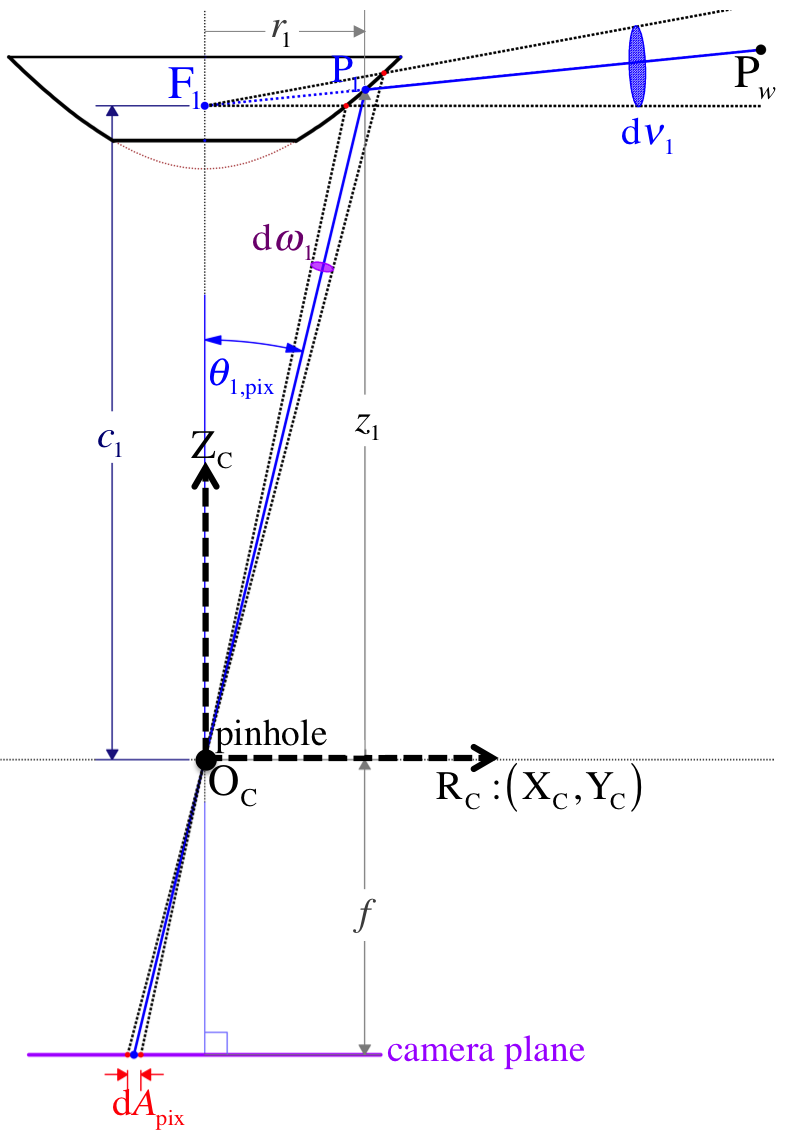}
    \caption[Spatial Resolution.]{The spatial resolution for the central catadioptric sensor
    is the ratio between a infinitesimal element of image area $\dif A$ and the corresponding solid
    angle infinitesimal element $\dif \nu_1$ that views the world point $\point{P}_w$. 
    (Notice that only mirror 1 is drawn in 2D and infinitesimal elements are exaggerated for easier
    visualization.)}
    \label{fig:spatial_resolution}
\end{figure} 

The resolution of the images acquired by our system are not space invariant. 
In fact, an omnidirectional camera producing spatial resolution-invariant images
can only be obtained through a non-analytical function of the mirror profile as shown in
\cite{Gaspar2002a}. In this section, we study the effect our design has on its spatial resolution as
it depends on position parameters like $d$ and $c_i$ introduced in \Fref{sec:model_parameters} as
well as a direct dependency on the characteristics (e.g. focal length $f$) of the camera obtaining
the image.

Let $\eta_{cam}$ be the spatial resolution for a conventional perspective camera as defined by Baker
and Nayar in \cite{Baker1999} and \cite{Swaminathan2001}. It measures the ratio between the
infinitesimal solid angle $\dif \omega_i$ (usually measured in steradians) that is directed toward a
point $\point{P}_i$ at an angle $\theta_{i,\mathrm{pix}}$ (formed with the optical axis
$\point{Z_C}$) and the infinitesimal element of image area $\dif A_\mathrm{pix}$ that $\dif
\omega_i$ subtends (as shown in \Fref{fig:spatial_resolution}). Accordingly, we have:
\begin{align}
    \label{eq:resolution_of_camera}
    \eta_{cam} &= 
    \frac{\dif A_\mathrm{pix}}{\dif \omega_i} = \frac{f^2}{\cos^3 \theta_{i,\mathrm{pix}}}
\end{align}
whose behavior 
tends to decrease as $\theta_{\mathrm{pix}} \to 0$, so higher resolution areas on the sensor plane
continuously increase the farther away they get from the optical center indicated at
$\cfs{\pixelPoint{m}_c}{}{I}$.
For ease of visualization, we plot only the $u$ pixel coordinates corresponding to the 2D spatial
resolution $\eta_{\mathrm{2D}}$, which is obtained by projecting the solid angle
$\mathnormal{\Omega}$ onto a planar angle $\theta_{\mathnormal{\Omega}}$ 
(the apex 2D angle of the solid cone of view). This yields
$\theta_{\mathnormal{\Omega}} = 2 \acos{1 - \mathnormal{\Omega} / {2 \pi}}$, and
we reduce the image area into its circular diameter with $2\sqrt{\dif A/\pi}$. Generally, our
conversion from 3D spatial resolution $\eta$ in $\left[\si{\square\metre}/\si{\steradian}\right]$ units to 2D proceeds as follows:
\begin{align}
    \label{eq:resolution_in_2D_conversion}
    \eta_{\mathrm{2D}} = \frac{2\sqrt{\eta/\pi}}{\theta_{\mathnormal{\Omega}=\SI{1}{\steradian}}}
\end{align}
where
$\theta_{\mathnormal{\Omega}=\SI{1}{\steradian}} \approx \SI{1.14390752211}{\radian}$.

More specifically, \fref{eq:resolution_of_camera} is manipulated to provide $\eta_{i,cam}$ as
the indicative of spatial resolution toward any specific point in the mirror,
$\cfs{\point{P}_i}{}{C} \in M_i$ according to \fref{eq:set_of_points_on_the_mirror}, as follows:
\begin{align}
    \label{eq:resolution_of_camera_toward_mirror_point}
    \eta_{i,cam}
    &= 
      \begin{cases}
        f^2 \left(\frac{ \sqrt{r_1^2 + z_1^2}}{z_1}\right)^3
        &\text{ if $i=1$,}
        \\    
        f^2 \left(\frac{\sqrt{r_2^2 + (d-z_2)^2}}{d - z_2}\right)^3
        &\text{ if $i=2$.}
      \end{cases}
\end{align}
where $r_i$ is the radial length defined in \fref{eq:r_limits} and its associated $z_i$ coordinate,
$f$ is the camera's focal length, and the design parameters $d$ and $c_i$ that relate to
the position of the mirror focal points $\point{F}_i$ with respect to the camera frame $\fr{C}$. 

Thus, for a conventional perspective camera, $\eta_{i,cam}$ grows
as $\theta_{i,\mathrm{pix}} \to \pi/2$ due to the
foreshortening effect that stretches the image representation around the sensor plane's periphery
that collects spatial information onto a larger number of pixels. Therefore, image areas
farther from the optical axis are considered to have higher spatial resolutions.

Baker and Nayar also defined the resolution, $\eta_{i}$, of a catadioptric
sensor in order to quantify the view of the world or $\dif
\nu_i$, an infinitesimal element of the solid angle subtended by the mirror's effective viewpoint
$\point{F}_i$, which is consequently imaged onto a pixel area $\dif A_\mathrm{pix}$. Again, here we
provide the resolution according to our model:
\begin{subequations}
    \label{eq:spatial_resolution_12}
    \begin{align}
        \label{eq:spatial_resolution_1}
        \eta_{1} &= \frac{\dif A_\mathrm{pix}}{\dif \nu_1}
        = \left[\frac{r_1^2 + (c_1 - z_1)^2)}{r_1^2 + z_1^2}\right] \eta_{1,cam}
        \\
        \label{eq:spatial_resolution_2}
        \eta_{2} &= \frac{\dif A_\mathrm{pix}}{\dif \nu_2}
        = \left[\frac{r_2^2 + (c_2 - d  + z_2)^2)}
        {r_2^2 + (d-z_2)^2}\right]\eta_{2,cam}
    \end{align}
\end{subequations}
for our mirror-perspective camera configuration, where $\point{O_C}$ is the origin of coordinates as
shown in \Fref{fig:spatial_resolution} and $\eta_{i,cam}$ is given in
\fref{eq:resolution_of_camera_toward_mirror_point}.

As demonstrated by the plot of
\Fref{fig:spatial_resolution_catadioptric_sensor} in \Fref{sec:spatial_resolution_optimality},
$\eta_{i}$ grows accordingly towards the periphery of each mirror (the equatorial region). This
aspect of our sensor design is very important because it indicates that the common field of view,
$\alpha_{SROI}$, where stereo vision is employed (\Fref{sec:3D_sensing}), is imaged at
relatively higher resolution than the unused polar regions closer to the optical axis (the
$\point{Z_C}$ axis).

If we modify $\eta_i$ by substituting $r_i$ with its equivalent
$\func{f}_{r_i}(z_i)$ function defined in \eqref{eq:func_r}, using mirror 1 for example, we
get:
\begin{align}
    \label{eq:spatial_resolution_alternative}
    \begin{split}
    \eta_{1} 
    &= \left[ \frac{\func{f}_{r_1}^2(z_1) + (c_1 - z_1)^2}{\func{f}_{r_1}^2(z_1) +
    z_1^2}\right]\eta_{1,cam}
    \\
    &= \frac
    {f^2 \sqrt{\func{f}_{r_1}^2(z_1) + z_1^2}
    \left[\func{f}_{r_1}^2(z_1) + (c_1 - z_1)^2\right]}
    {z_1^3}
    \end{split}
\end{align}
which is an inherent indicative of how the resolution $\eta_{i}$ for a reflection point
$\point{P}_i$ increases with $k \to \infty$, as plotted in \Fref{fig:spatial_resolution_of_mirrors_for_various_k_values}.
Conversely, the smaller the $k_i$ parameter gets (as it
relates to eccentricity, see \Fref{sec:SVP_configuration}), the flatter the mirror becomes, so its
resolution resembles more that of the perspective camera alone, or mathematically
$
\lim_{k_i \to 2} \eta_i \to \eta_{i,cam}
$.

As shown in \Fref{fig:effect_of_k_on_system_radius}, a smaller $k_i$ would require a wider radius
$r_{sys}$ in order to achieve the same omnidirectional vertical field of view, $\alpha_{sys}$. Even worse, in
order to image such a wider reflector, either the camera's field of view, $\alpha_{cam}$, would have
to increase (by decreasing the focal length $f$ and perhaps requiring a larger camera hole $r_{cam}$
and sensor size), or the distance $c_i$ between the effective pinhole and the viewpoint would have
to increase accordingly.
Another consequence, it is the effect on the baseline $b$, which must change in order to
maintain the same vertical field of views (\Fref{fig:effect_of_k1_on_baseline}).
As a result, the depth resolution of the stereo system would suffer as indicated next.

\section{Parameter Optimization and Prototyping}
\label{sec:optimization_and_prototyping}

The nonlinear nature of this system makes it very difficult to balance among its desirable
performance aspects.
The optimal vector of design parameters, $\vecttheta^*$, can be found by posing a constrained
maximization problem for the objective function 
\begin{align}
    \label{eq:baseline_function}
    \func{f}_b(\vecttheta) = c_1 + c_2 - d
\end{align}
which
measures the baseline according to \fref{eq:baseline}. Indeed, the optimization problem is subject to
the set of constraints $C$, which we enumerate in \Fref{sec:optimization_constraints}.
Formally,
\begin{align}
\label{eq:optimization_of_baseline}
    \vecttheta^* = \argmax_{\vecttheta \in \Theta} \func{f}_b(\vecttheta) ~ \text{ subject to } C
\end{align}
where $\Theta\subseteq \mathbb{R}^6$ is the 6-dimensional solution space for $\vecttheta \in
\mathbb{R}^6$ given in \eqref{eq:vector_of_primary_design_parameters} as $\vecttheta
=\begin{bmatrix}c_1,&c_2,&k_1,&k_2,&d,&r_{sys}\end{bmatrix}$.

\subsection{Optimization Constraints}
\label{sec:optimization_constraints}

This section discusses the constraints that the proposed omnistereo sensor is subject to. Overall,
we mainly take the following into account:
\begin{enumerate}[a)]
  \item \emph{geometrical constraints}, including SVP and reflex constraint, as described by
  equations \eqref{eq:hyperboloid1}, \eqref{eq:hyperboloid2}, and \eqref{eq:reflex_plane};
  \item \emph{physical constraints}, these are the rig's dimensions, which include the mirrors radii
  as well as by-product parameters such as system height $h_{sys}$ and mass
  $m_{sys}$;
  \item \emph{performance constraints}, where the spatial resolution and range from triangulation
  are determined by parameters $k_1$, $k_2$, and $c_1$, and the desired viewing angles
  for an optimal Stereo ROI and the related common field of view, $\alpha_{SROI}$.
\end{enumerate}

Following the design model described throughout
\Fref{sec:sensor_design}, we now list the pertaining linear and nonlinear
constraints that make the set $C$. We disjoint the linear constraints in a subset $C_L$ so those
non-linear constraints belong to another subset $C_{NL}$, and $C = C_L \uplus C_{NL}$. Within each
subset, we generalize equality constraints as functions $\func{h}\colon \mathbb{R}^6 \mapsto
\mathbb{R}$ that obey
\begin{align}
    \label{eq:equality_constraint_rule}
    \func{h}(\vecttheta) = 0
\end{align}
whereas inequality functions $\func{g}\colon \mathbb{R}^6 \mapsto
\mathbb{R}$ satisfy
\begin{align}
    \label{eq:inequality_constraint_rule}
    \func{g}(\vecttheta) \le 0
\end{align}

\subsubsection{Linear Constraints}
\label{sec:linear_constraints}

We have only setup linear inequalities for constraints in $C_L$. Specifically, we require the
following:

\begin{enumerate}[$\func{g}_1$:]
    \item  The focal distance $c_2$ of mirror 2 must be longer than $d$ (distance between
    $\point{O_C}$ and $\point{F}_{2v}$). Requiring that 
    \begin{align}
        \label{eq:constraint_linear_ineq1}
        d \le c2
    \end{align}
    in order to set the position of 
    $\point{F}_2$ below the origin $\point{O_C}$ of the pinhole camera frame $\fr{C}$.
    \item
    Mirror 1's focal separation, $c_1$, needs to exceed the placement of the reflex mirror. 
    By letting 
        \begin{align}
            \label{eq:constraint_linear_ineq2}
            d/2  \le c_1
        \end{align}
    the hyperboloidal mirror can reflect light towards its effective
    viewpoint $\point{F}_1$ without being occluded by the reflex mirror.
    \item The empirical constraint 
    \begin{align}
        \label{eq:constraint_linear_ineq3}
        \dfrac{5}{3} \le \dfrac{k_2}{k_1}
    \end{align}
    pertains our rig
    dimensions in order to assign a greater curvature to mirror 2's profile (located a the bottom), so it can look
    more toward the equatorial region rather than up.
    Complementarily, this constraint flattens mirror 1's profile, so it can possess a greater view
    of the ground rather than the sky. This curvature inequality allows the Stereo ROI to be
    bounded by a wider vertical field of view when the sensor must be mounted above the MAV
    propellers as depicted in \Fref{fig:FOVs}.
\end{enumerate}

\subsubsection{Non-Linear Constraints}
\label{sec:nonlinear_constraints}

For the non-linear design constraints, we have the following inequalities:

\begin{enumerate}[$\func{g}_1$:]
    \setcounter{enumi}{3}
    \item The AscTec Pelican quadrotor has a maximum payload of \SI{650}{\gram}
    (according to the manufacturer specifications \cite{AscTec2014}). Therefore, we
    must satisfy the system mass' equation computed via \eqref{eq:mass_sys}, such that
    \begin{align}
        \label{eq:constraint_nonlinear_ineq1}
        m_{sys} \le 650
    \end{align}
    \item
    Similarly, we limit the system's height \fref{eq:height} by
    $h_{sys,max}$.
        \begin{align}
            \label{eq:constraint_nonlinear_ineq2}
            h_{sys} \le h_{sys,max}
        \end{align}
    For example, we set $h_{sys,max} = \SI{150}{\milli\metre}$ for the
    \SI{37}{\milli\metre}-radius rig.
    \item 
        The origin of coordinates for the camera frame is set at it viewpoint, $\point{O_C}$.
        In order to fit the camera enclosure under mirror 2, it is realistic to position the
        vertex on the vertical transverse axis at more than \SI{5}{\milli\metre} away from $\point{O_C}$:
        \begin{align}
            \label{eq:constraint_nonlinear_ineq3}
            5 \le z_{0_2} - a_2
        \end{align}
        where $z_{0_2}$ is defined in \eqref{eq:z0_2}, and $a_2$ pertains to \eqref{eq:hyperboloid}.
\end{enumerate}

Next, we determine the bounds for the limiting angles that partake in the
computation of the system's vertical field of view $\alpha_{sys}$, which is based on equation
\eqref{eq:vFOV_system}.
Our application has specific viewing requirements that can be achieved with the following
conditions:
\begin{enumerate}[$\func{g}_1$:]
    \setcounter{enumi}{6}
    \item  Let $\Lambda_{1,max}=\SI{14}{\degree}$ be an acceptable upper-bound for angle 
    $\theta_{1,max}$ 
    , such that:
    \begin{align}
        \label{eq:constraint_nonlinear_ineq4}
        \theta_{1,max} \le \Lambda_{1,max}
    \end{align}
    \item Because we desire a larger view towards the ground from mirror 1, we empirically set
    $\Lambda_{1,min}=\SI{-25}{\degree}$ as a lower-bound for the minimum elevation
    $\theta_{1,min}$ 
:
    \begin{align}
        \label{eq:constraint_nonlinear_ineq5}
        \Lambda_{1,min} \le \theta_{1,min} 
    \end{align}
    \item In order to avoid occlusions with the MAV's propellers while being capable to image
    objects located about \SI{5}{\centi\metre} under the rig's base and \SI{20}{\centi\metre} away
    from the central axis, we limit mirror 2's lowest angle by a lower-bound
    $\Lambda_{2,min}=\SI{-14}{\degree}$.
    Symbolically, 
    \begin{align}
        \label{eq:constraint_nonlinear_ineq6}
        \Lambda_{2,min} \le \theta_{2,min}
    \end{align}
\end{enumerate}

Finally, we restrict the radius of the system, $r_{sys}$, to be identical for both hypeboloids by
satisfying the following equality condition:

\begin{enumerate}[$\func{h}_1$:]
    \item  \begin{par}
        With functions $\func{f}_{r_1}$ and $\func{f}_{r_2}$ defined in
        \eqref{eq:func_r}, we set
        \begin{align*}
            r_{sys} = r_{i,max} &= \func{f}_{r_i}(z_{i,max}), \forall i\in\{1,2\}         
        \end{align*}
        where we imply that $z_{i,max} \gets \func{f}_{z_i}(r_{sys})$ using
        \fref{eq:func_z}. Thus, the entire function composition for this equality becomes
        \begin{align}
            \label{eq:constraint_nonlinear_equality}
            \func{f}_{r_1}\left(\func{f}_{z_1}(r_{sys})\right) &=
            \func{f}_{r_2}\left(\func{f}_{z_2}(r_{sys})\right)
        \end{align}
    \end{par}
\end{enumerate}



\subsection{Optimal Results}
\label{sec:prototype_dimensions}

\begin{figure}
    \centering
    \includegraphics[width=\columnwidth,keepaspectratio]{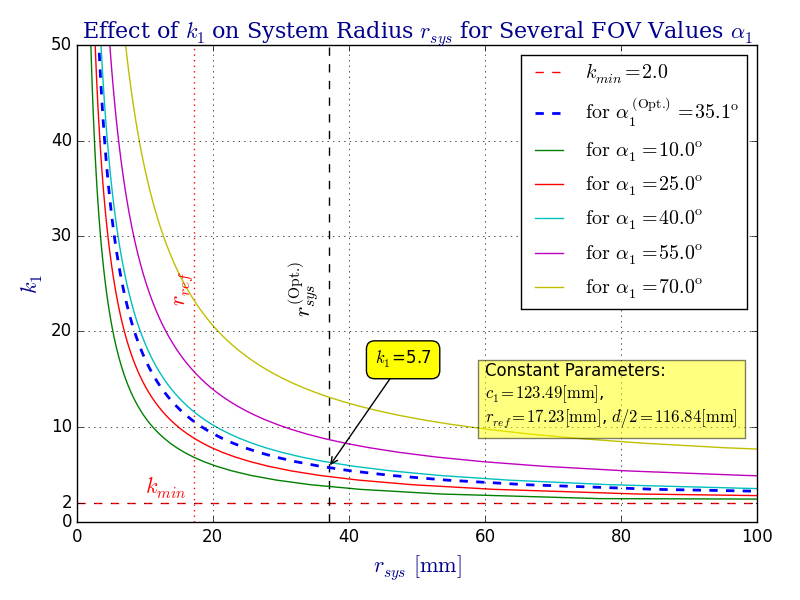}
    \caption[The effect of k on the system radius]{The effect that parameter $k_i$ (showing mirror 1
    only) has over the system radius $r_{sys}$ for various vertical field of
    view angles $\alpha_1$. In order to maintain a vertical field of
    view $\alpha_i$ that is bounded by $z_{max}\mid_{r_{sys}}$, the value of $r_{sys}$ must
    change accordingly.
    Inherently,
    the system's height, $h_{sys}$, and its mass, $m_{sys}$, are also affected by $k_i$
    (see \Fref{sec:size}).
    }
    \label{fig:effect_of_k_on_system_radius}
\end{figure}

\begin{figure}
    \centering
    \includegraphics[width=\columnwidth,keepaspectratio]{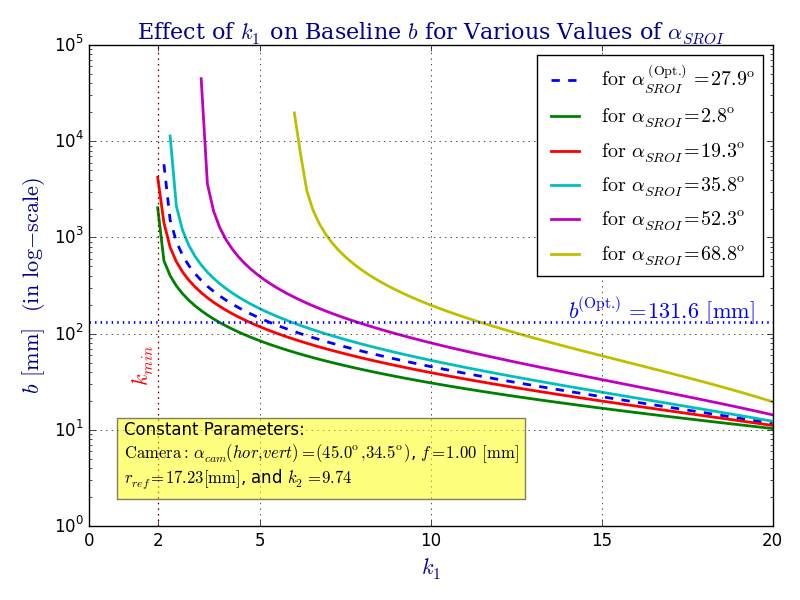}
    \caption[The effect of k1 on the system baseline]{The effect that parameter $k_1$ has over the
    omnistereo system's baseline $b$ for several common FOV angles ($\alpha_{SROI}$) and a fixed
    camera with $\alpha_{cam}$.
    An inverse relationship exists between $k$ and $b$ as plotted here (using a logarithmic scale for the
    vertical axis). Intuitively, the flatter the mirror gets ($k \to 2$), the farther $\point{F}_1$
    must be translated in order to fit within the camera's view, $\alpha_{SROI}$, causing $b$
    to increase.
    }
    \label{fig:effect_of_k1_on_baseline}
\end{figure}

\begin{figure*}
    \centering
    \includegraphics[width=1.0\textwidth,keepaspectratio]{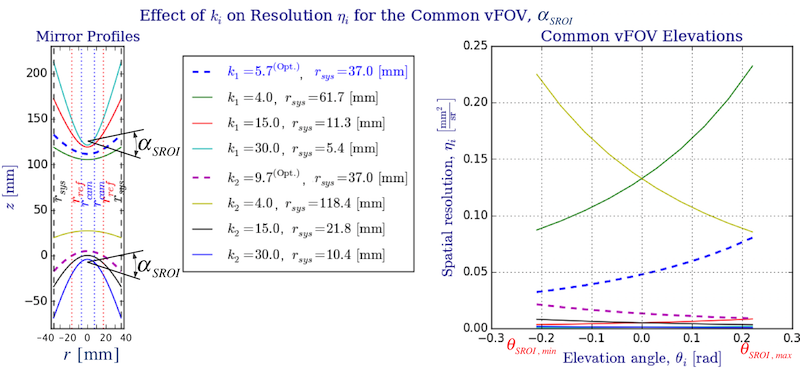}
    \caption[Spatial Resolutions for various value of k.]{Comparison of $k_i$ values
    and their effect on spatial resolution $\eta_i$. Note that for the big rig, the optimal focal dimensions
    $c_1$ and $c_2$ (from \Fref{tab:optimal_parameter_values}) were used as well as the angular span
    on the common vertical FOV, $\alpha_{SROI} \approx \ang{28}$.
    Although resolution $\eta_{i}^{(Opt.)}$ for the optimal values of $k_i$ could be improved by
    employing smaller $k$ values (lower curvature profiles indicated on the left plot of the
    figure), these would in turn increase the system radius, $r_{sys}$, so they are not desirable.
    As expected, we appreciate how the spatial resolutions, $\eta_i$ for $i=\{1,2\}$, increase
    towards the equatorial regions ($\theta_1 \to \theta_{SROI,max}$ and $\theta_2 \to
    \theta_{SROI,min}$).
    }
    \label{fig:spatial_resolution_of_mirrors_for_various_k_values}
\end{figure*} 

Applying the aforementioned constraints (\Fref{sec:optimization_constraints}) and using an iterative
nonlinear optimization method such as one of the 
surveyed in \cite{Forsgren2002}, a bounded solution vector $\vecttheta^*$ converges to the
following values for two rig sizes:
\begin{table}[H]
  \caption{Optimal System Design Parameters}
  \label{tab:optimal_parameter_values}
  \begin{center}
    \begin{tabular}
      {| c || S | S | }
        \hhline{|---|}
        Parameter & {Big Rig} &  {Small Rig} \\
        \hhline{|=#=|=|} 
        $b = \max \func{f}_b(\vecttheta^*)$ & 131.61 & 108.92
        \\
        \hhline{|---|} 
        $r_{sys} [\si{\milli\metre}]$   &  37.0 & 28.0
        \\
        $c_1 [\si{\milli\metre}]$ & 123.49 & 104.59
        \\
        $c_2 [\si{\milli\metre}]$ & 241.80 & 204.34
        \\
        $d [\si{\milli\metre}]$ &  233.68 & 200.00
        \\ 
        $k_1$ & 5.73 & 6.88
        \\
        $k_2$ & 9.74 & 11.47
        \\
        \hhline{|---|}
      \end{tabular}
    \end{center}
\end{table}

\begin{table}
    \caption{By-product Length Parameters}
    \label{tab:byproduct_length_parameters}
    \begin{center}
    \begin{tabular}
      {| c || S | S | }
        \hhline{|---|}
        Parameter & {Big Rig} &  {Small Rig} \\
        \hhline{|=#=|=|} 
        $r_{ref} [\si{\milli\metre}]$ & 17.23 & 11.74
        \\
        $r_{cam} [\si{\milli\metre}]$ & 7 & 7
        \\
        $h_{sys} [\si{\milli\metre}]$ & 150.00 & 120.00
        \\
        \hhline{|---|}
      \end{tabular}
      \end{center}
\end{table}

As \Fref{fig:geometric_model} illustrates, a realistic dimension for the radius of the camera hole,
$r_{cam}$, must consider the maximum value between a physical micro-lens radius ($r_{lens}$) and the
radius $r_{\alpha_{cam}}$ for an unoccluded field of view of the camera ($\alpha_{cam}$).
Practically,
\begin{equation}
    \label{eq:r_cam}
    r_{cam} = \max \left(r_{lens},  r_{\alpha_{cam}} \right)
\end{equation}
\Fref{tab:byproduct_length_parameters} contains the derived dimensions corresponding to the
parameters listed in \Fref{tab:optimal_parameter_values}.

For both rigs, the expected vertical field of views are
$\alpha_{sys}=\ang{75}-(\ang{-21}) \approx \ang{96}$ according to \fref{eq:vFOV_system}, and
$\alpha_{SROI}=\ang{14}-(\ang{-14}) \approx \ang{28}$ using \fref{eq:vFOV_common}.
Note that $\theta_{2,max}$ may be actually limited by the camera hole radius, which in turn reduces
$\theta_{cam} \leadsto \ang{59}$, and the real $\alpha_{sys} \leadsto \ang{80}$.
For the big rig, \Fref{tab:vertices_stereo_ROI} shows the nearest vertices of the Stereo ROI
that result from these angles (\Fref{fig:common_FOV_ROI}).

\begin{table}
    \caption{Near Vertices of the Stereo ROI for the Big Rig}
    \label{tab:vertices_stereo_ROI}
    \begin{center}
    \begin{tabular}
      {| c || S | S | }
        \hhline{|---|}
        Vertex & {$\cfs{\rho_w}{}{C}$ [\si{\milli\metre}]} &  {$\cfs{z_w}{}{C}$ [\si{\milli\metre}]}
        \\
        \hhline{|=#=|=|} 
        $\point{P}_{ns_{high}}$ & 93.5 & 144.4
        \\
        $\point{P}_{ns_{mid}}$ & 65.2 & 98.4
        \\
        $\point{P}_{ns_{low}}$ & 763.4 & -170.3
        \\
        \hhline{|---|}
      \end{tabular}
      \end{center}
\end{table}

Finally, we study the effect parameter $k_i$ has over the system
radius $r_{sys}$ (\Fref{fig:effect_of_k_on_system_radius}), the omnistereo baseline $b$
(\Fref{fig:effect_of_k1_on_baseline}), and the spatial resolution
(\Fref{fig:spatial_resolution_catadioptric_sensor}
and \Fref{fig:spatial_resolution_of_mirrors_for_various_k_values}).
\Fref{fig:effect_of_k_on_system_radius} addresses the relation between
$k_i$ and radius $r_{sys}$ (recall the rig size discussion from
\Fref{sec:size}). 
It can be seen that for the same $r_{sys}$, realistic $k_1$ values fall in the range $3 < k_1 < 13$,
whereas the vertical field of view $\alpha_1 \to 0$ as $k\to 2$ as expected according to the SVP
property specified in \Fref{sec:SVP_configuration}.
In fact, the left part of \Fref{fig:spatial_resolution_of_mirrors_for_various_k_values} also
demonstrates the necessary $r_{sys}$ to maintain $\alpha_{SROI} \approx \ang{28}$
for various values of $k_i$.

\Fref{fig:effect_of_k1_on_baseline} shows the inverse relationship
between values of $k_1$ and the baseline, $b$, as we attempt to fit the view of a wider/narrower
mirror profile (due to $k_1$) on the constant camera field of view ($\alpha_{cam}$). 
In order to make a fair comparison, let 
\[
k_1'= k_1+\varepsilon_k, \quad \forall k_1 > 2, \varepsilon_k > 0
\] 
for which we find its new focal length $c_1'$ while solving for the new $r_{sys}'$ and
$z_{max}'$. Provided with a function such that $c_1 \gets \func{f}_{c_1}(k_1)$,
we perform the analysis for a given $\alpha_{SROI}$ and
$\alpha_{cam}$ shown in \Fref{fig:effect_of_k1_on_baseline}.
Given the baseline function $\func{f}_b$ defined in \eqref{eq:baseline_function}, the following
implication holds true:
\begin{align}
    \label{eq:baseline_change_due_to_k}
    \func{f}_b\mid_{c_1 \gets \func{f}_{c_1}(k_1)} >
    \func{f}_b\mid_{c_1'\gets \func{f}_{c_1}(k_1+\varepsilon_k)},
    ~
    \forall k_1 > 2, \varepsilon_k > 0
\end{align}
Notice that $k_2$, $c_2$ and $d$ are kept constant through this analysis, and we ignore possible
occlusions from the reflex mirror located at $d/2$.

\subsubsection{Spatial Resolution Optimality}
\label{sec:spatial_resolution_optimality}

\begin{figure}
    \centering
    \includegraphics[width=\columnwidth,keepaspectratio]{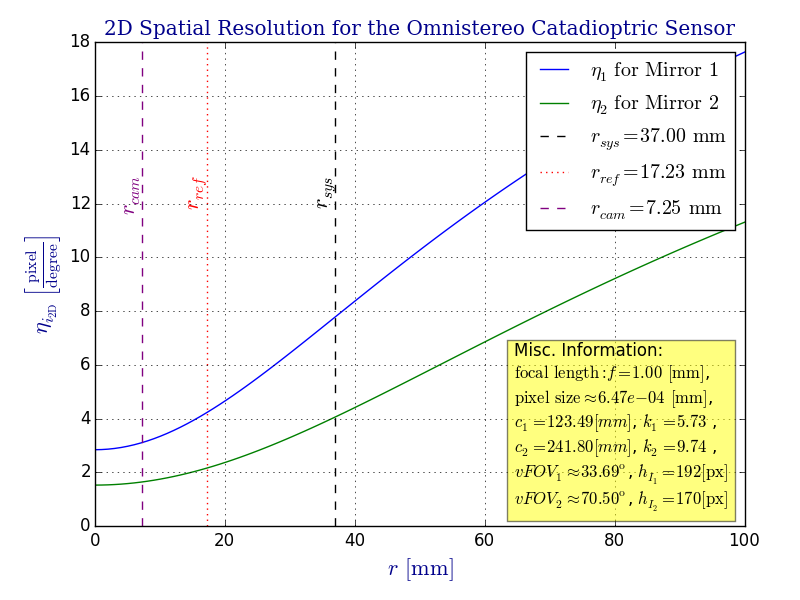}
    \caption[Spatial Resolution of the Catadioptric Sensor.]{Using the formula given in
    \eqref{eq:resolution_in_2D_conversion}, we plot the 2D version of the spatial resolution of our
    proposed omnistereo catadioptric sensor (\SI{37}{\milli\metre}-radius rig). We observe
    how both resolutions $\eta_1$ and $\eta_2$ increase towards the equatorial region where they are
    physically limited by $r_{sys}$. In turn, this verifies the spatial resolution theory given in
    \cite{Baker1999}, and it justifies our coaxial configuration useful for omnistereo within the
    Stereo ROI indicated in \Fref{fig:common_FOV_ROI}.}
    \label{fig:spatial_resolution_catadioptric_sensor}
\end{figure}

In this section, we compare the sensor's spatial resolution, $\eta_{i}$, defined in
\Fref{sec:spatial_resolution} for the optimal parameters listed in \Fref{tab:optimal_parameter_values}  (for the big
rig, only). In \Fref{fig:spatial_resolution_catadioptric_sensor}, we verify
how both resolutions $\eta_1$ and $\eta_2$ increase towards the equatorial region
according to the spatial resolution theory presented in
\cite{Baker1999}. Indeed, the increase in spatial resolution within the Stereo ROI that covers
the equatorial region (as indicated in \Fref{fig:common_FOV_ROI}) justifies our model's
coaxial configuration intended as an omnistereo application.

In \Fref{fig:spatial_resolution_of_mirrors_for_various_k_values}, we compare the
effect on $\eta_i$ for various mirror profiles, which depend directly on $k_i$. We illustrate the
change in curvature due to parameter $k_1$ and $k_2$ and also show (in the legend) the respective
$r_{sys}$ achieving a common vFOV, $\alpha_{SROI} \approx \ang{28}$ as for the optimal parameter of
the big rig. From this plot, we appreciate the compromise due to optimal parameters,
$k_1^{(Opt.)} = 5.7$ and $k_2^{(Opt.)} = 9.7$, for a balanced system size due to $r_{sys}$ and a
suitable range of spatial resolutions, $\eta_i$, within the Stereo ROI enveloped by
$\alpha_{SROI}$.

For the optimal parameter values
listed in \Fref{tab:optimal_parameter_values}, we find that $\chi_{\point{I}_{1:2}} \approx 2$.

\subsection{Prototypes}
\label{sec:prototypes}

We validate our design with both synthetic and real-life models.

\subsubsection{Synthetic Prototype (Simulation)}
\label{sec:synthetic_prototype}

After converging to an optimal solution $\func{f}_b(\vecttheta^*)$, we employ these parameters
(\Fref{tab:optimal_parameter_values}) to describe synthetic models using POV-Ray, an open-source
ray-tracer. We render 3D scenes via the simulated omnistereo sensor such as the examples shown in
\Fref{fig:sim_office}.
The simulation stage plays two important roles in our investigation:
\begin{enumerate}[1)]
    \item to acquire ground-truth 3D-scene information in order to evaluate the computed range by
    the omnistereo system (as explained in \Fref{sec:3D_sensing}, and 
    \item to provide an almost
    accurate geometrical representation of the model by discounting some real-life computer vision
    artifacts such as assembly misalignments, glare from the support tube (motivates use
    of standoffs on real prototype), as well as the shallow camera's depth-of-field.
    All of these can affect the quality of the real-life results shown in this manuscript
    (\Fref{sec:3D_sensing_results}).
\end{enumerate} 

\subsubsection{Real-life Prototypes}
\label{sec:prototype_real}
\begin{figure}
     \captionsetup[subfloat]{justification=centering}
         \begin{subfigure}[t]{0.5\textwidth}
                 \centering
                 \vspace{0pt} 
                 \includegraphics[height=55mm,keepaspectratio]{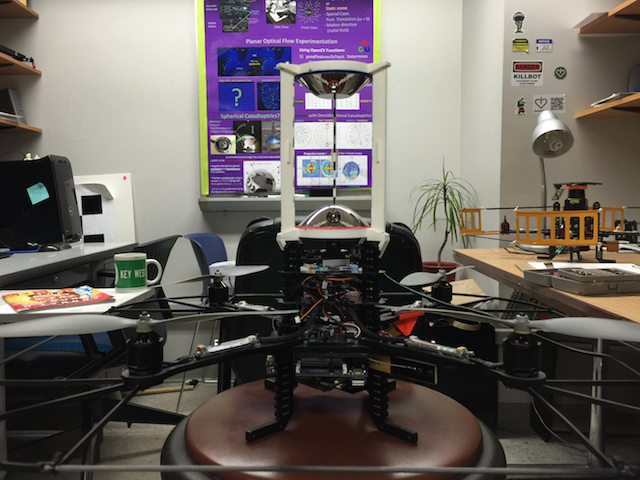}
                 \caption{Omnistereo rig using \SI{37}{\milli\metre}-radius mirrors mounted on
                 an AscTec Pelican quadrotor.}
                 \label{fig:74mm_prototype_on_pelican}
         \end{subfigure}%
         \hfill
         \begin{subfigure}[t]{0.4\textwidth}
                 \centering
                 \vspace{0pt} 
                 \includegraphics[height=55mm,keepaspectratio]{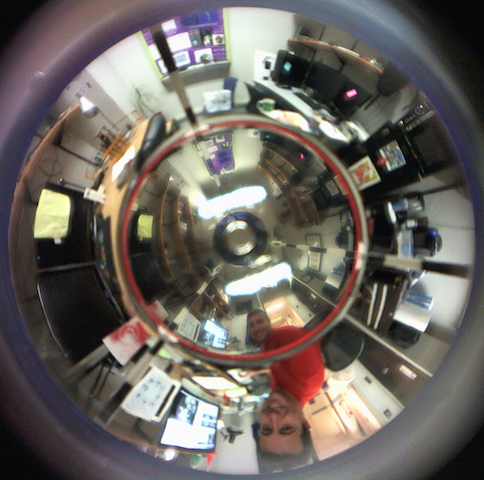}
                 \caption{Omnidirectional image captured by 
            the real-life prototype in
             \Fref{fig:74mm_prototype_on_pelican}.}
                 \label{fig:74mm_prototype_camera_view}
         \end{subfigure}
      \clearcaptionsetup{subfloat}
         \caption[Real-life prototypes]{Real-life prototypes of the omnistereo sensor.
         rig.}%
         \label{fig:prototyped_rigs}%
\end{figure}

We have also produced two physical prototypes
that can be installed on the Pelican quadrotor (made by Ascending Technologies). 
A small rig
assembled with
hyperboloidal mirrors of $r_{sys} \approx \SI{28}{\milli\metre}$ (according to the specifications
from \Fref{tab:optimal_parameter_values}). 

\Fref{fig:74mm_prototype_on_pelican} shows the larger rig constructed with
hyperboloidal mirrors of $r_{sys} \approx \SI{37}{\milli\metre}$,
and a Logitech\textsuperscript{\textregistered} HD Pro Webcam C910 camera (2592x1944 pixels
at $15 \sim 20$ FPS).
We decided to skip the use of acrylic glass tubes to separate the mirrors at the specified
$h_{sys}$ distance, and instead we constructed a thin 3-standoff mount in order to avoid glare and
cross-reflections. This support was designed in 3D-CAD and
printed for assembly. The three areas of occlusion due to the \SI{3}{\milli\metre}-wide standoffs are considered minimal. 
In fact, we stamped fiducial markers to the vertical standoffs to aid
with the panoramas generation (\Fref{sec:panoramic_images}) and auto-calibration in the
future.
To image the entire surface of mirror 1, we require a camera with a (minimum) field of view of
$\alpha_{cam} > \ang{31}$, which is achieved by $r_{\alpha_{cam}} > \SI{1.4}{\milli\metre}$.
In practice, as noted by \fref{eq:r_cam}, microlenses measure around
$r_{lens} \approx \SI{7}{\milli\metre}$.
Therefore, we set $r_{cam} > \SI{7}{\milli\metre}$, as a safe specification to fit a standard
microlens through the opening of mirror 2 as shown in \Fref{fig:geometric_model}.

Recall that $m_{sys}$
is limited by the maximum
\SI{650}{\gram}-payload that the AscTec Pelican quadrotor is capable of flying
with (according to the manufacturer specifications \cite{AscTec2014}).
The camera with lens weights approximately \SI{25}{\gram}. The cylindrical tube material is
acrylic with average density $\rho_{tub} \approx \SI{1.18}{\gram\per\cubic\centi\metre}$, whereas
the brass that is used to mold the mirrors has a density
$\rho_{mir} \approx \SI{8.5}{\gram\per\cubic\centi\metre}$. 
Empirically, we obtain a close estimate of the entire system's mass, such that
$m_{sys}\approx \SI{550}{\gram}$ for the big rig, and $m_{sys}\approx \SI{150}{\gram}$ for the small
rig.

\section{3D Sensing from Omnistereo Images}
\label{sec:3D_sensing}

Stereo vision from point correspondences on images at distinct locations is a popular method for
obtaining 3D range information via triangulation. Techniques for image point matching 
are generally divided between dense (area-based scanning \cite{Forsgren2002}) and sparse (feature
description \cite{Tuytelaars2008}) approaches.
Due to parallax, the disparity in point positions for objects
close to the vision system is larger than for objects that are farther away. As illustrated in
\Fref{fig:common_FOV_ROI}, the nearsightedness of the sensor is determined mainly by the
common observable space (a.k.a. Stereo ROI) acquired by the limiting elevation angles of the
mirrors (\Fref{sec:FOV}). In addition, we will see next (\Fref{sec:triangulation}) that the baseline
$b$ also plays a major role in range computation.

Due to our model's coaxial configuration, we could scan for pixel correspondences radially between
a given pair of warped images $\left(\fr{I_1},\fr{I_2}\right)$ like in the
approach taken by similar works such as \cite{Spacek2004}. However, it seems more convenient to work on a
rectified image space, such as with panoramic images, where the search for correspondences can be
performed using any of the various existing methods for perspective stereo views. Hence, we first
demonstrate how these rectified panoramic images are produced (\Fref{sec:panoramic_images}) and used
for establishing point correspondences. Then, we proceed to study our triangulation method for
the range computation from given set of point correspondences (\Fref{sec:triangulation}). Last, we
show preliminary 3D point clouds as the outcome from such procedure.

\subsection{Panoramic Images}
\label{sec:panoramic_images}

\begin{figure}
    \centering
    \includegraphics[width=\columnwidth,keepaspectratio]{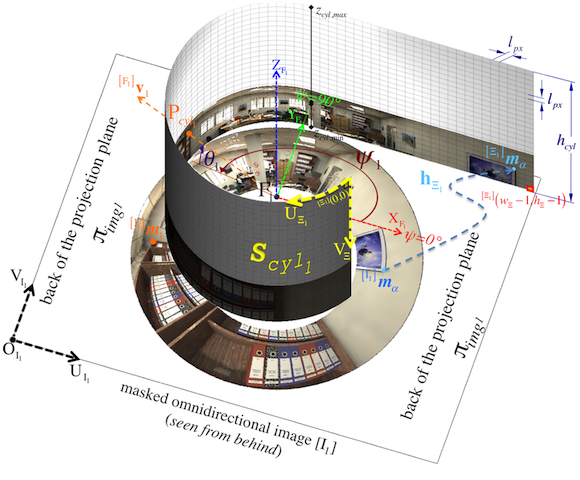}
    \caption[Panoramic Image Formation]{An example for the formation of panoramic image
    $\fr{\Xi_1}$ out of the omnidirectional image $\fr{I_1}$ (showing only the masked region of
    interest on the back of image plane $\pi_{img_1}$).
    Any particular ray, $\vect{v}_1$
    indicated by its elevation and azimuth such as
    $\cfs{\left(\psi_1,\theta_1 \right)}{}{F_1}$ that is directed
    towards the focus $\point{F_1}$ must traverse the projection cylinder $S_{cyl_1}$ at point
    $\point{P}_{cyl_1}$.
    More abstractly, the figure also shows how a
    pixel position $\cfs{\pixelPoint{m}_\alpha}{}{\Xi_1}$ on the panoramic pixel space is mapped
    from its corresponding pixel position $\cfs{\pixelPoint{m}_\alpha}{}{I_1}$ via function
    $\funcvv{h}_{\Xi_1}$ defined in \eqref{eq:panorama_pixel_mapping_function}. Although not up to
    scale, it's crucial to notice the relative orientation between $S_{cyl_1}$ and the {\em back} of the
    projection plane $\pi_{img_1}$ where the omnidirectional image $\fr{I_1}$ is found.}
    \label{fig:panoramic_image_formation}
\end{figure}

\Fref{fig:panoramic_image_formation} illustrates how we form the respective
panoramic image $\fr{\Xi_1}$ out of its warped omnidirectional image $\fr{I_1}$. As illustrated in
\Fref{fig:imaging_ratio}, $\fr{I_i}$ is simply the region of interest out of the full image $\fr{I}$ where
projection occurs via mirror $i$. However, we can safely refer to $\fr{I}$ because it will
never be the case that projections via different mirrors overlap on the same pixel position $\cfs{\pixelPoint{m}}{}{I}$.
In a few words, we obtain a panorama $\fr{\Xi_i}$ by reverse-mapping each discretized 3D point
$\point{P}_{cyl_i} \in S_{cyl_i}$ to its projected pixel coordinates $\cfs{\pixelPoint{m}}{}{I}$ on
$\fr{I}$ according to \Fref{sec:back_projection}.

More thoroughly, for $i=\{1,2\}$, $S_{cyl_i}$  is the set of all valid 3D points $\point{P}_{cyl_i}$
that lie on an imaginary unit cylinder centered along the Z-axis and positioned with respect to the
mirror's primary focus $\point{F}_i$.
Recall that the radius of a unit cylinder is $r_{cyl} = 1$, so its circumference becomes
$w_{cyl} = 2 \pi r_{cyl} = 2 \pi$.
Noticed that the imaging ratio, $\chi_{\point{I}_{1:2}}=\frac{h_{I_1}}{h_{I_2}}$, illustrated in \Fref{fig:imaging_ratio} provides a way of inferring the scale between pairs of
point correspondences. However, we achieve conforming scales among both panoramic
representations by simply setting both cylinders to an equal height $h_{cyl}$, which is determined from
the system's elevation limits, $(\theta_{sys,min}, \theta_{sys,max})$, since they partake in the
measurement of the system's vertical field of view \eqref{eq:vFOV_system}.
Therefore, we obtain
\begin{align}
    \label{eq:cylinder_height}
    h_{cyl} = z_{cyl,max}  - z_{cyl,min} 
    ~\text{, where } \left\{ ~ 
    \begin{array}{ll}
    z_{cyl,max} &= \tan \left(\theta_{sys,max}\right)
    \\
    z_{cyl,min} &= \tan \left(\theta_{sys,min}\right)
    \end{array}
    \right.
\end{align}

Consequently, to achieve panoramic images $\fr{\Xi_i}$ of the same dimensions by maintaining a
true aspect ratio $w_{\Xi}:h_{\Xi}$, it suffices to indicate either the
width (number of columns) $w_{\Xi}$ or the height (number of rows) $h_{\Xi}$ as number of pixels.
Here, we propose a custom method for resolving the panoramic image dimensions by setting the
equality for the length $l_{px}$ of an individual ``square'' pixel in the cylinder (behaving
like a panoramic camera sensor):
\begin{align}
    \label{eq:panorama_pixel_size}
    l_{px}= \frac{w_{cyl}}{w_{\Xi}} = \frac{h_{cyl}}{h_{\Xi}}
\end{align}
If the width $w_{\Xi}$ is given, then the height is
given by $h_{\Xi} = w_{\Xi} h_{cyl} / w_{cyl}$. Otherwise, if $h_{\Xi}$ is given a priori, we can
find $w_{\Xi} = h_{\Xi} w_{cyl} / h_{cyl}$.

To increase the processing speed for each panoramic image $\fr{\Xi_i}$, we fill up its corresponding
look-up-table LUT$_{\Xi_i}$ of size $w_{\Xi} \times h_{\Xi}$ that encodes the mapping for each
panoramic pixel coordinates $\cfs{\pixelPoint{m}}{}{\Xi_i}=\cfs{\vectTwoColInline{u}{v}}{}{\Xi_i}$
to its respective projection $\cfs{\pixelPoint{m}}{}{I_i}=\cfs{\vectTwoColInline{u}{v}}{}{I_i}$.
Each pixel $\cfs{\pixelPoint{m}}{}{\Xi_i}$ gets associated with its cylinder's 3D point
positioned at $\cfs{\vect{p}_{cyl_i}}{}{F_i}$, which can inherently be indicated by its elevation
$\cfs{\theta_{i}}{}{F_i}$ and azimuth
$\cfs{\psi_{i}}{}{F_i}$ (relative to the mirror's primary focus
$\point{F}_i$) as illustrated in \Fref{fig:image_formation}. Thus, the ray
$\cfs{\vect{v}_i}{}{F_i}$ of a particular 3D point directed about $\cfs{\left(\psi_i, \theta_i \right)}{}{F_i}$ must pass through
$\point{P}_{cyl_i}$ in order to get imaged as pixel
$\cfs{\pixelPoint{m_i}}{}{I}$.

Since the circumference of the cylinder, $w_{cyl}$, is discretized with respect to the number of
pixel columns or width $w_{\Xi}$, we use the pixel length $l_{px}$ as the factor to obtain the
arc length $l_{\psi_{i}}$ spanned by the azimuth $\cfs{\psi_{i}}{}{F_i}$ out of
a given $\cfs{u}{}{\Xi_i}$ coordinate on the panoramic image. Generally,
\begin{align}
\label{eq:azimuth_inference_from_column_in_panoramic_image}
    \cfs{\psi_{i}}{}{F_i} = \frac{l_{\psi_{i}}}{r_{cyl}} = 
        \frac{w_{cyl} - \cfs{u}{}{\Xi_i}~l_{px}}{r_{cyl}}
\end{align}
or simply $\cfs{\psi_{i}}{}{F_i} = 2\pi - \cfs{u}{}{\Xi_i}~l_{px}$ for the unit cylinder case.

An order reversal in the columns of the
panorama is performed by \fref{eq:azimuth_inference_from_column_in_panoramic_image}
because we account for the relative position between $S_{cyl_i}$ and the projection plane
$\pi_{img}$.
For $\fr{\Xi_1}$, \Fref{fig:panoramic_image_formation} depicts the unrolling of the cylindrical
panoramic image onto a planar panoramic image. Recall that the projection to the image plane
However, note that $\pi_{img}$
is shown from above (or its back) in \Fref{fig:panoramic_image_formation}, so the panorama
visualization places the viewer inside the cylinder at $\point{F}_1$.

Similarly, the elevation angle $\cfs{\theta_{i}}{}{F_i}$ is inferred out the row or
$\cfs{v}{}{\Xi_i}$ coordinate, which is scaled to its cylindrical representation by
$l_{px}$. Recall that both cylinders have the same height, $h_{cyl}$, computed by
\fref{eq:cylinder_height}, and by taking into account any row offset from the top or maximum height
position, $\cfs{z_{cyl,max}}{}{F_i}$, of the cylinder, we get
\begin{align}
\label{eq:elevation_inference_from_row_in_panoramic_image}
    \cfs{\theta_{i}}{}{F_i} &= \atan{\cfs{z_{cyl,max}}{}{F_i} - \cfs{v}{}{\Xi_i}~l_{px}} 
\end{align}

Assuming coaxial alignment, out of these angles, we evaluate the positon vector
$\cfs{\vect{p}_{cyl_i}}{}{C}$ for a point on the panoramic cylinder with respect to the camera frame
$\fr{C}$:
\begin{align}
\label{eq:point_in_cylindrical_panorama}
    \cfs{\vect{p}_{cyl_i}}{}{C} &= 
    \cancel{r_{cyl}} \begin{bmatrix}
    \cos \left( \cfs{\psi_{i}}{}{F_i} \right)
    \\
    \sin \left( \cfs{\psi_{i}}{}{F_i} \right)
    \\
    \tan \left( \cfs{\theta_{i}}{}{F_i} \right)
    \end{bmatrix}
    + \cfs{\vect{f}_i}{}{C}
\end{align}
where $r_{cyl}$ cancels out for a unit cylinder.

Therefore, we use the direction equations 
\eqref{eq:azimuth_inference_from_column_in_panoramic_image} and
\eqref{eq:elevation_inference_from_row_in_panoramic_image} leading to
\eqref{eq:point_in_cylindrical_panorama} as a process:
$\cfs{\pixelPoint{m}}{}{\Xi_i} 
    \xmapsto[{
    \text{
    \eqref{eq:elevation_inference_from_row_in_panoramic_image}
    }
    }
    ]
    {\text{
    \eqref{eq:azimuth_inference_from_column_in_panoramic_image}
    }
    }
    \cfs{
    \begin{pmatrix}
    \psi \\
    \theta
    \end{pmatrix}
    }{}{F_i}
    \xmapsto{\text{\eqref{eq:point_in_cylindrical_panorama}}}
    \cfs{\vect{p}_{cyl_i}}{}{C}
$, which is eventually used as the input argument to
\fref{eq:forward_projection_function} in order to determine pixel $\cfs{\pixelPoint{m}}{}{I_i}$
via the mapping function $\funcvv{h}_{\Xi_i}: \mathbb{R}^2 \mapsto \mathbb{R}^2$
as following:
\begin{align}
    \label{eq:panorama_pixel_mapping_function}
    \cfs{\pixelPoint{m}}{}{I_i} \gets \funcvv{h}_{\Xi_i}(\cfs{\pixelPoint{m}}{}{\Xi_i})
    &\colonequals \funcvv{f}_{\varphi_i}\left( 
    \left.\cfs{\vect{p}_{cyl_i}}{}{C} \right|_{\negthickspace
    \scalebox{0.6}{\cfs{\pixelPoint{m}}{}{\Xi_i}} } \right) 
\end{align}

\subsubsection{Stereo Matching on Panoramas}
\label{sec:stereo_matching}

\begin{figure*}
    \centering
    \includegraphics[width=\textwidth,keepaspectratio]{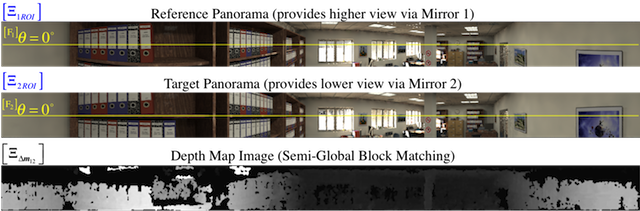}
    \caption[Dense Stereo Matching on Panoramic Images]{
    For the synthetic omnidirectional image $\fr{I}$ shown in \Fref{fig:sim_rig_camera_view}, we
    generate its pair of panoramic images
    $\left(\fr{\Xi_1},\fr{\Xi_2}\right)$ using the procedure explained in
    \Fref{sec:panoramic_images}. Note that we only work on the Stereo ROI (shown here) to perform a
    semi-global block match between the panoramas as indicated in \Fref{sec:stereo_matching}. 
    The resulting disparity map, $\fr{\Xi_{\Delta m_{12}}}$, is visualized at the bottom as a
    gray-scale panoramic image normalized about its 256 intensity levels, where brighter colors
    imply larger disparity values. To aid the relative vertical view of both panoramas,
    we have annotated the row position of the zero-elevation. }
    \label{fig:panoramic_dense_stereo_matching}
\end{figure*}

We understand that the algorithm chosen for finding matches is crucial to attain correct pixel
disparity results.
We refer the reader to \cite{Bradski2008} for a detailed description of stereo correspondence
methods.
After comparing various block matching algorithms, we were able to obtain acceptable disparity maps
with the semi-global block matching (SGBM) method introduced by \cite{Hirschmuller2008}, which
can find subpixel matches in real time. As a result of this stereo block matcher among
the pair of panoramic images $\left(\fr{\Xi_1},\fr{\Xi_2}\right)$, we get the
dense disparity map $\fr{\Xi_{\Delta m_{12}}}$ visualized as an
image in \Fref{fig:panoramic_dense_stereo_matching}. Note that valid disparity values must be positive
$(>0)$ and they are given with respect to the reference image, in this case, $\fr{\Xi_1}$.
In addition, recall that no stereo matching algorithm (as far as we are aware) is totally immune to
mismatches due to several well-known reasons in the literature such as ambiguity of cyclic
patterns.

An advantage of the block (window) search for
correspondences is that it can be narrowed along epipolar lines. Unlike the traditional horizontal
stereo configuration, our system captures panoramic images whose views differ in a vertical fashion.
As shown in \cite{Gluckman1998a}, the unwrapped panoramas contain vertical, parallel epipolar lines
that facilitate the pixel correlation search. Thus, given a pixel position
$\cfs{\pixelPoint{m}_1}{}{\Xi_1}$ on the reference panorama $\fr{\Xi_1}$ and its
disparity value $\left. \Delta m_{12}\right
|_{\negthickspace\scalebox{0.6}{\cfs{\pixelPoint{m}_1}{}{\Xi_1}}}$, we can resolve the
correspondence $\cfs{\pixelPoint{m}_2}{}{\Xi_2}$ pixel coordinate on the target image, $\fr{\Xi_2}$,
by simply offsetting the $v$-coordinate with the disparity value:
\begin{align}
    \label{eq:correspondence_from_vertical_block_match}
    \cfs{\pixelPoint{m}_2}{}{\Xi_2} &= 
        \begin{bmatrix}
        u_1
        \\
        v_1 + \left. \Xi_{\Delta m_{12}} \right
        |_{\negthickspace\scalebox{0.6}{\cfs{\pixelPoint{m}_1}{}{\Xi_1}}}
        \end{bmatrix}
\end{align}

\subsection{Range from Triangulation}
\label{sec:triangulation}

Recall the duality that states a point $\point{P}_w$ as the intersection of a pair of lines.
Regardless of the correspondence search technique employed, such as block stereo matching
between panoramas $\fr{\Xi_i}$  (\Fref{sec:stereo_matching}) or feature detection directly on
$\fr{I}$, we can resolve for $\cfs{\left(\pixelPoint{m}_1, \pixelPoint{m}_2\right)}{}{I}$.
From equations \eqref{eq:bp_ray1_wrt_F1} and
\eqref{eq:bp_ray2_wrt_F2}, we obtain the respective pair of back-projected rays
$\left(\cfs{\vect{v}_1}{}{F_1}, \cfs{\vect{v}_2}{}{F_2}\right)$, emanating from their respective
physical viewpoints, $\point{F}_1$ and $\point{F}_2$, which are separated by baseline $b$.
We can compute elevation angles $\theta_1$ and $\theta_2$ using equations
\eqref{eq:theta1} and \eqref{eq:theta2}. Then, we can triangulate the back-projected rays in order
to calculate the horizontal range $\rho_w$ defined in \eqref{eq:external_position_constraint},
as follows:
\begin{equation}
    \label{eq:depth_from_triangulation}
    \rho_w = \abs{\frac{{b\cos({\theta _1})\cos({\theta _2})}}{{\sin ({\theta_1} -
    {\theta _2})}}}
\end{equation} 
Finally, we obtain the 3D position of $\point{P}_w$:
\begin{equation}
    \label{eq:pw_from_triangulation}
    \cfs{\vect{p}_w}{}{C} = \begin{bmatrix}
        -\rho_w \cos (\psi_{12} ) \\
        -\rho_w \sin (\psi_{12} ) \\
        c_1 - \rho_w \tan(\theta_1)
    \end{bmatrix}
\end{equation}
where $\psi_{12}$ is the common azimuthal angle (on the XY-plane) for coplanar rays, so it can be
determined either by \fref{eq:azimuth1} or \eqref{eq:azimuth2}. 
Functionally, we define the ``naive'' intersection function that implements
\fref{eq:pw_from_triangulation} and \eqref{eq:depth_from_triangulation} such that
\begin{align}
    \label{eq:function_triangulation}
    \cfs{\vect{p}_w}{}{C} \gets \funcvv{f}_{\Delta}((\theta_1,\psi_1), (\theta_2,
    \psi_2),\vecttheta)
\end{align}
where $\vecttheta$ is the model parameters vector defined in
\fref{eq:vector_of_primary_design_parameters} and can be omitted when calling this function because
the model parameters should not change (ideally).

\begin{figure}
    \centering
    \includegraphics[width=0.6\textwidth,keepaspectratio]{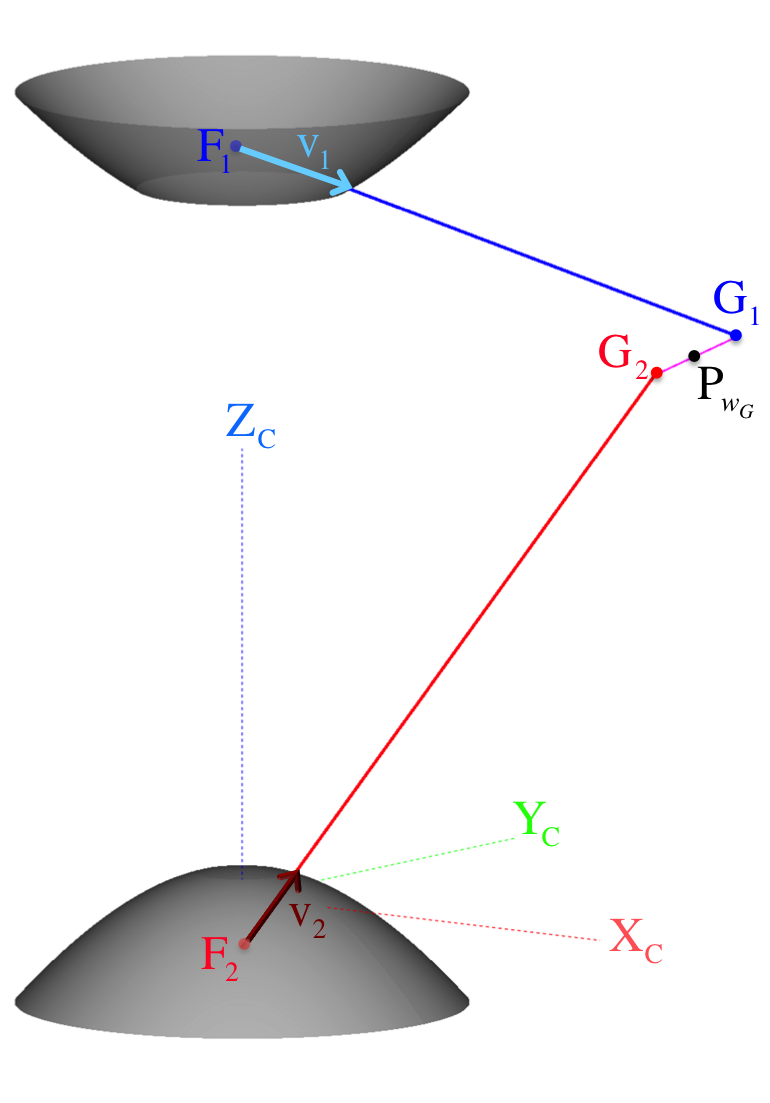}
    \caption[Triangulation with skew back-projection rays.]{The more realistic case of
    skew back-projection rays $(\vect{v}_1,\vect{v}_2)$ approximates the triangulated point
    $\point{P}_w$ by getting the midpoint $\point{P}_{w_G}$ on the common perpendicular line segment 
    $\overline{\point{G}_1\point{G}_2}: \lambda_{1 \perp 2} \vect{\hat{v}}_{1 \perp2}$. Note that
    these skey rays were formed from a pixel correspondence pair $\cfs{(\pixelPoint{m}_1,
    \pixelPoint{m}_2)}{}{I}$ and by offsetting the coordinate $v_2$ by 15 pixels.
    }
    \label{fig:triangulation_skew_rays}
\end{figure}

\subsubsection{Common Perpendicular Midpoint Triangulation Method}
\label{sec:common_perpendicular_midpoint_triangulation_method}
Because the coplanarity of these rays cannot be guaranteed (skew rays case), a better
triangulation approximation while considering coaxial misalignments is to find the midpoint of their
common perpendicular line segment (as attempted in \cite{He2008}).
As illustrated in \Fref{fig:triangulation_skew_rays}, we define the
common perpendicular line segment $\overline{\point{G}_1\point{G}_2}$ as the
parametrized vector $\vect{v}_{1 \perp 2} = \lambda_{1 \perp 2} \vect{\hat{v}}_{1 \perp
2}$, for the unit vector normal to the back-projected rays, $\vect{v}_1$ and $\vect{v}_2$, such
that:
\begin{align}
    \label{eq:common_perpedicular_unit_vector}
    \vect{\hat{v}}_{1 \perp 2} = \frac{\vect{v}_1 \otimes \vect{v}_2}
    {\mynorm*{\vect{v}_1 \otimes \vect{v}_2}}
\end{align}

If the rays are not parallel ($\mynorm{\vect{v}_1 \otimes \vect{v}_2} \ne 0$), we can compute the
``exact'' solution,
$\vect{\lambda} = \vectThreeColInline{\lambda_{G_1}}{\lambda_{G_2}}{\lambda_{1 \perp 2}}$,
of the well-determined linear matrix equation
\begin{align}
    \label{eq:linear_system_common_perpendicular}
    \matx{V}\vect{\lambda} &= \vect{b}
    ~\text{, where} ~
    \matx{V} = \begin{bmatrix}
                \vect{v}_1, && -\vect{v}_2, && \vect{\hat{v}}_{1 \perp 2}
                \end{bmatrix}
    \text{ and }
    \vect{b} = \cfs{\vect{f}_2}{}{C} - \cfs{\vect{f}_1}{}{C}
\end{align}
It follows that the location of the midpoint $\point{P}_{w_G}$ on the common perpendicular
$\vect{v}_{1 \perp 2}$ with respect to the common frame $\fr{C}$ is
\begin{align}
    \label{eq:midpoint_on_common_perpendicular}
    \cfs{\vect{p}_{w_G}}{}{C} &= \cfs{\vect{f}_1}{}{C} 
    + \lambda_{G_1}\cfs{\vect{v}_1}{}{F_1}
    + \frac{1}{2}\lambda_{1 \perp 2} \cfs{\vect{\hat{v}}_{1 \perp2}}{}{G_1}
\end{align} 

\subsubsection{Range Variation}
\label{sec:range_variation_comparisson}

\begin{figure}
    \centering
    \includegraphics[width=\columnwidth,keepaspectratio]{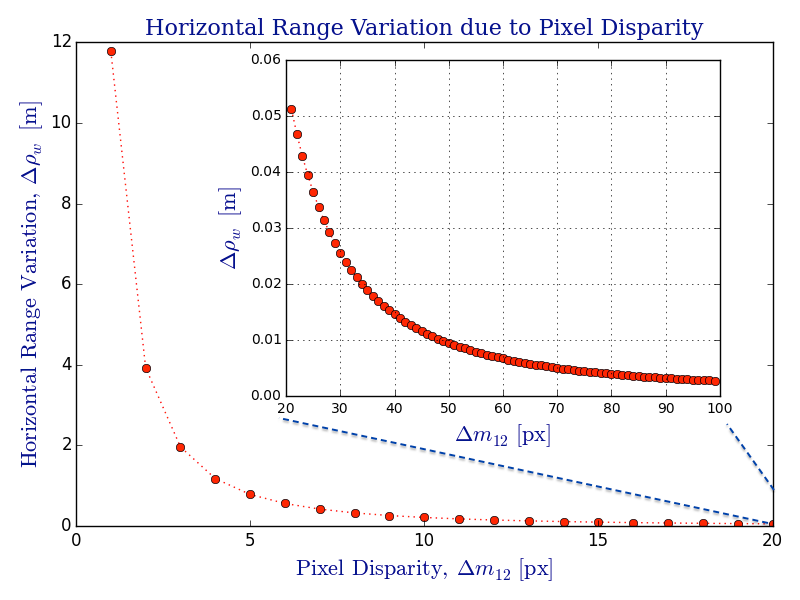}
    \caption[Horizontal range variation due to pixel disparity]{Variation of
    horizontal range, $\Delta \rho_w$, due to change in pixel disparity $\Delta m_{12}$ on the
    omnidirectional image, $\fr{I}$.
    There exists a ``nonlinear \& inverse'' relation between the change in depth from triangulation
    ($\Delta \rho_w$) and the number of disparity pixels ($\Delta m_{12}$) available from the
    omnistereo image pair $\left(\fr{I_1},\fr{I_2}\right)$, which are exclusive subspaces of
    $\fr{I}$.
    }
    \label{fig:horizontal_range_variation_due_to_pixel_disparity}
\end{figure}

Before we introduce an uncertainty model for triangulation
(\Fref{sec:triangulation_uncertainty_model}), we briefly analyze how range varies according to the
possible combinations of pixel correspondences, $\cfs{(\pixelPoint{m}_1, \pixelPoint{m}_2)}{}{I}$
on the image $\fr{I}$.
Here, we demonstrate how a radial variation of discretized pixel disparities, $\Delta m_{12}$,
affects the 3D position of a point obtained from triangulation (\Fref{sec:triangulation}). 
\Fref{fig:horizontal_range_variation_due_to_pixel_disparity} demonstrates the
nonlinear characteristics of the variation in horizontal range, $\Delta \rho_w$, from the discrete
relation between pixel positions $\cfs{\pixelPoint{m}_i}{}{I}$ and their respective back-projected (direction) rays
obtained from $\funcvv{f}_{\beta_i}$ and triangulated via function $\funcvv{f}_{\Delta}$ defined in
\eqref{eq:function_triangulation}.
It can be observed that the horizontal range variation, $\Delta \rho_w$, increases quadratically as
$\Delta m_{12} \to 1 \mathrm{px}$, which is the minimum {\em discrete} pixel disparity providing
a maximum horizontal range $\rho_{w,max} \approx \left[18,28\right]$ \si{\metre} (computed
analytically).
The main plot of \Fref{fig:horizontal_range_variation_due_to_pixel_disparity} shows the small
disparity values in the interval $\overline{\Delta m_{12}} =
[1,20]~\mathrm{px}$, whereas the subplot is a zoomed-in extension of the large disparity cases in
the interval $\overline{\Delta m_{12}} = [20,100]~\mathrm{px}$. 

The current analysis is an indicative that triangulation
error (e.g. due to false pixel correspondences) may have a severe effect on range accuracy that
increases quadratically with distance as it can be appreciated with the \SI{8}{\metre} variation
on the disparity interval $\overline{\Delta m_{12}} = [1,2] \mathrm{px}$
Also, observe the example of \Fref{fig:point_cloud_synthetic_example} for a reconstructed point
cloud, where this range sensing characteristic is more noticeable for faraway points. 
In fact, the
following uncertainty model provides a probabilistic framework for the triangulation error
(uncertainty) that agrees with the current numerical claims.

\subsection{Triangulation Uncertainty Model}
\label{sec:triangulation_uncertainty_model}

\begin{figure}
    \centering
    \includegraphics[width=0.4\textwidth,keepaspectratio]{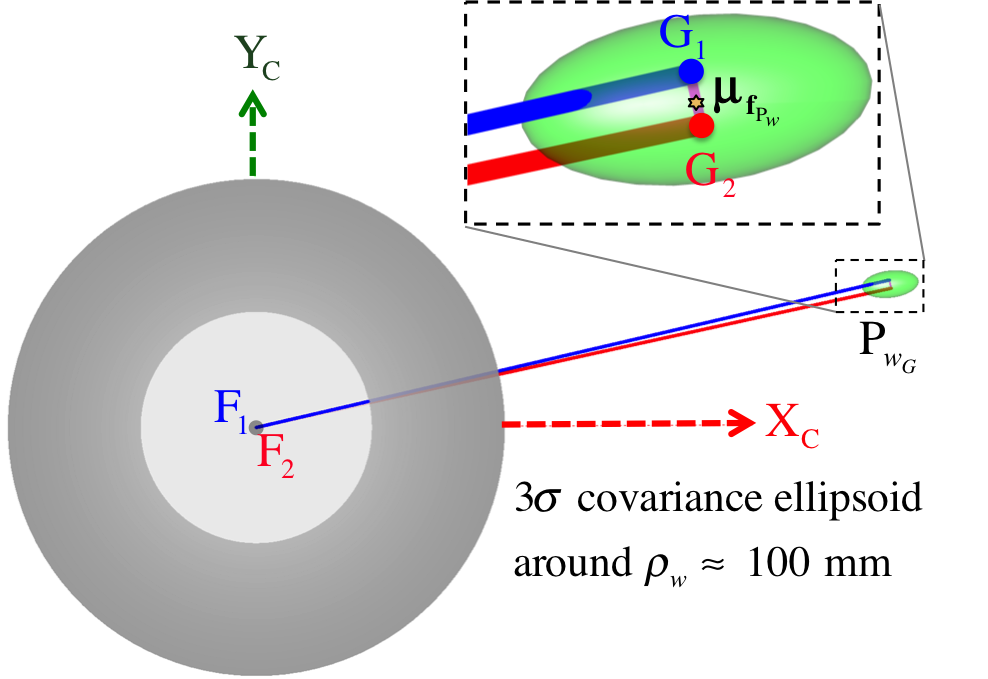}
    \caption[Visualization of Uncertainty Ellipsoid for 1px error]{Top-view of the
    three-sigma level ellipsoid for the triangulation uncertainty of a pixel pair
    $\cfs{(\pixelPoint{m}_1, \pixelPoint{m}_2)}{}{I}$ with an assumed standard deviation
    $\sigma_{px}=1~\mathrm{px}$. Observe how this
    $1~\mathrm{px}$ deviation skews the back-projected rays, so the midpoint $\point{P}_{w_G}$
    on the common perpendicular line segment $\overline{\point{G}_1\point{G}_2}$ is employed to
    estimate the mean position $\cfs{\vectmu_{\funcvv{f}_{\point{P}_w}}}{}{C}$ with $\rho_w \approx
    \SI{100}{\milli\metre}$.
    }
    \label{fig:uncertainty_ellipsoid_for_1px_skew-top_view}
\end{figure}

\begin{figure*}
    \centering
    \includegraphics[width=\textwidth,keepaspectratio]{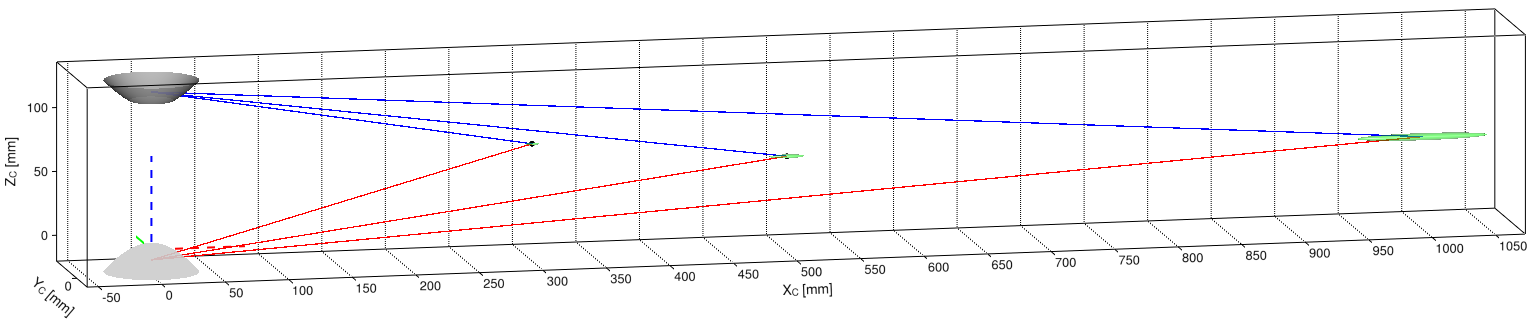}
    \caption[Visualization of Uncertainty Ellipsoids (Covariances) from Triangulation]{Visualization
    of a few triangulated points at mean positions $ \cfs{\vectmu_{\funcvv{f}_{\point{P}_w}}}{}{C}$
    using horizontal ranges: $\rho_w \approx \{0.3, 0.5, 1.0\}$ \si{\metre} and their
    uncertainty/error ellipsoids representing their corresponding covariance matrix $\bm{\Sigma}_{\funcvv{f}_{\point{P}_w}}$ (drawn for
    one-$\sigma_{\funcvv{f}_{\point{P}_w}}$ level and assuming $\sigma_{px}=1~\mathrm{px}$). Here,
    it is evident how the triangulation uncertainty dominates along the outward-radial direction of
    horizontal range $\rho_w$.}
    \label{fig:uncertainty_ellipsoids}
\end{figure*}

Let $\funcvv{f}_{\point{P}_w}$ be the vector-valued function that computes the 3D coordinates of
point $\point{P}_{w_G}$ with respect to $\fr{C}$ as the common perpendicular midpoint defined in \fref{eq:midpoint_on_common_perpendicular}.
We express this triangulation function component-wise as follows:
\begin{align}
    \label{eq:function_triangulation_midpoint}
    \cfs{\vect{p}_{w_G}}{}{C} \gets \funcvv{f}_{\point{P}_w}(\vect{m}_{12}) \colonequals
    \vectThreeRow{\func{f}_{x_w}(\vect{m}_{12})}{\func{f}_{y_w}(\vect{m}_{12})}{\func{f}_{z_w}(\vect{m}_{12})}
\end{align}
where $\vect{m}_{12}=[u_1, v_1, u_2, v_2]$ is composed by the pixel coordinates of the correspondence
 $\cfs{(\pixelPoint{m}_1, \pixelPoint{m}_2)}{}{I}$ upon which to base the
triangulation (\Fref{sec:triangulation}).

Without loss of generality, we model a {\em multivariate
Gaussian} uncertainty model for triangulation, so that the position vector
$\cfs{\vect{p}_{w_G}}{}{C}$ of any world point is centered at its mean
$\cfs{\vectmu_{\funcvv{f}_{\point{P}_w}}}{}{C}$ with a $3\times3$ covariance matrix
$\bm{\Sigma}_{\funcvv{f}_{\point{P}_w}}$:
\begin{align}
    \label{eq:gaussian_uncertainty_model_for_triangulation}
    \cfs{\vectmu_{\funcvv{f}_{\point{P}_w}}}{}{C} = \vectThreeRow{x_{w}}{y_{w}}{z_{w}},  
    &~
   \bm{\Sigma}_{\funcvv{f}_{\point{P}_w}} = 
    \begin{bmatrix}
    \sigma_{\func{f}_{x_w}}^2 
    && \sigma_{\func{f}_{x_w}}\sigma_{\func{f}_{y_w}} 
    && \sigma_{\func{f}_{x_w}}\sigma_{\func{f}_{z_w}}
    \\
    \sigma_{\func{f}_{x_w}}\sigma_{\func{f}_{y_w}} 
    && \sigma_{\func{f}_{y_w}}^2 
    && \sigma_{\func{f}_{y_w}}\sigma_{\func{f}_{z_w}}
    \\
    \sigma_{\func{f}_{x_w}}\sigma_{\func{f}_{z_w}} 
    && \sigma_{\func{f}_{y_w}}\sigma_{\func{f}_{z_w}} 
    && \sigma_{\func{f}_{z_w}}^2
    \end{bmatrix}
\end{align}

However, since $\funcvv{f}_{\point{P}_w}$ is a non-linear vector-valued function, we linearize it by
approximation to a first-order Taylor expansion and we use its Jacobian matrix to propagate the
uncertainty (covariance) as in the linear case as follows:
\begin{align}
    \label{eq:uncertainty_propagation}
   \bm{\Sigma}_{\funcvv{f}_{\point{P}_w}} = 
    \matx{J}_{\funcvv{f}_{\point{P}_w}}
   ~ \bm{\Omega}_{\vect{m}_{12}} ~
    \trans{\matx{J}_{\funcvv{f}_{\point{P}_w}}}
\end{align}
where the $3\times4$ Jacobian matrix for the triangulation
function is
\begin{align}
    \label{eq:jacobian_matrix_triangulation_function}
    \matx{J}_{\funcvv{f}_{\point{P}_w}} = 
    \begin{bmatrix}
        \frac{\partial\func{f}_{x_w}}{\partial u_1}
        &
        \frac{\partial\func{f}_{x_w}}{\partial v_1}
        &
        \frac{\partial\func{f}_{x_w}}{\partial u_2}
        &
        \frac{\partial\func{f}_{x_w}}{\partial v_2}
        \\
        \frac{\partial\func{f}_{y_w}}{\partial u_1}
        &
        \frac{\partial\func{f}_{y_w}}{\partial v_1}
        &
        \frac{\partial\func{f}_{y_w}}{\partial u_2}
        &
        \frac{\partial\func{f}_{y_w}}{\partial v_2}
        \\
        \frac{\partial\func{f}_{z_w}}{\partial u_1}
        &
        \frac{\partial\func{f}_{z_w}}{\partial v_1}
        &
        \frac{\partial\func{f}_{z_w}}{\partial u_2}
        &
        \frac{\partial\func{f}_{z_w}}{\partial v_2}
    \end{bmatrix}
\end{align}
and the $4x4$ covariance matrix of the pixel arguments being 
\begin{align}
    \label{eq:pixel_covariance_matrix}
   \bm{\Omega}_{\vect{m}_{12}} = \sigma_{px}^2\matx{I}_4 
\end{align}
where we assume $\sigma_{px}=1~\mathrm{px}$ for the standard deviation of each
pixel coordinate in the {\em discretized} pixel space.
The complete symbolic solution of $\bm{\Sigma}_{\funcvv{f}_{\point{P}_w}}$ is too involved to appear in this manuscript. 
However, in \Fref{fig:uncertainty_ellipsoid_for_1px_skew-top_view}, we show the top-view of the
covariance ellipsoid drawn at a three-$\sigma_{\funcvv{f}_{\point{P}_w}}$ level for a point
triangulated nearly around $\rho_w \approx \SI{100}{\milli\metre}$.
\Fref{fig:uncertainty_ellipsoids} visualizes uncertainty ellipsoids drawn at a
one-$\sigma_{\funcvv{f}_{\point{P}_w}}$ level for several triangulation ranges.
We refer the reader to the end of \Fref{sec:sparse_triangulation_rmse} where we validate this safe
assumption through experimental results using subpixel precision.

\section{Experiment Results and Future Work}
\label{sec:experiments_and_future_work}

\subsection{3D Sensing Results}
\label{sec:3D_sensing_results}

\begin{figure*} 
    \begin{subfigure}[c]{0.6\textwidth}
            \includegraphics[height=50mm,keepaspectratio]{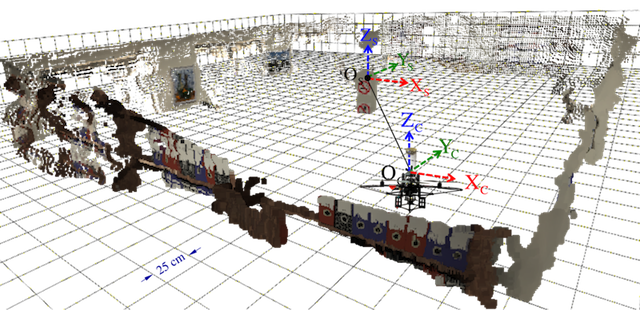}
            \subcaption{3D Perspective View}
            \label{fig:point_cloud_synthetic_example_3D_view}
    \end{subfigure}
    \hfill
    \begin{subfigure}[c]{0.30\textwidth}
            \includegraphics[height=50mm,keepaspectratio]{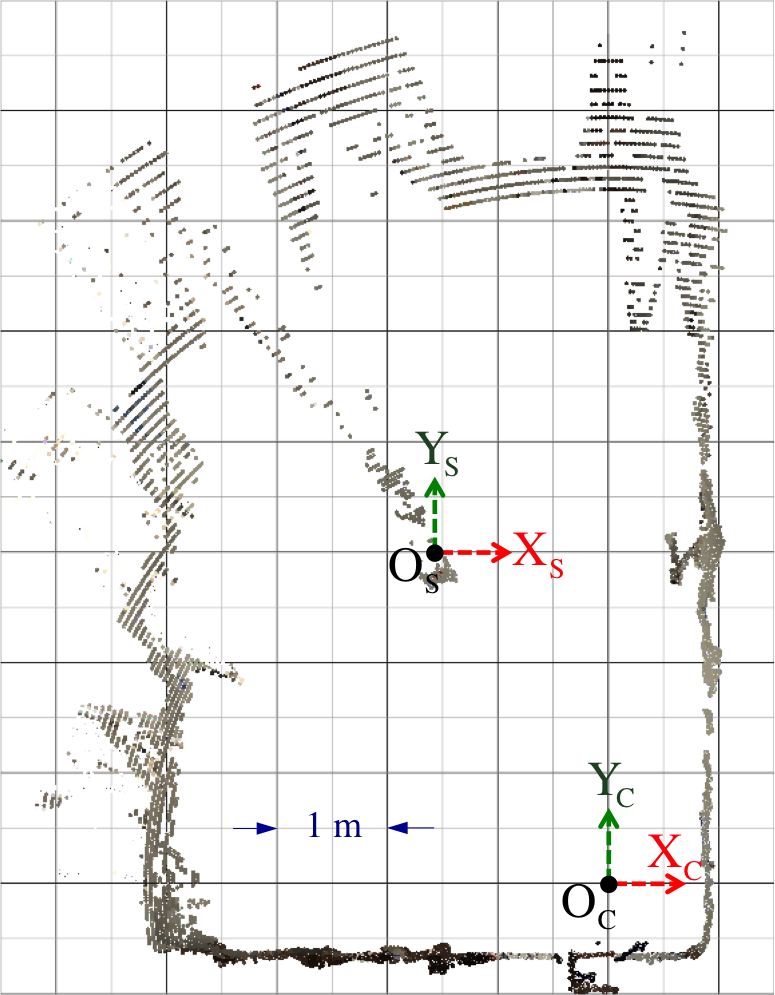}
            \subcaption{Orthographic View}
            \label{fig:point_cloud_synthetic_example_ortho_view}
    \end{subfigure}
    
    \caption[3D Dense Point Cloud Example from Synthetic Model]{A 3-D dense point cloud computed
    out of the synthetic model that rendered the omnidirectional image shown in
    \Fref{fig:sim_rig_camera_view}. Pixel correspondences are established via the 
    panoramic depth map visualized in \Fref{fig:panoramic_dense_stereo_matching}. The
    3D point triangulation implements the common perpendicular midpoint method indicated in
    \Fref{sec:common_perpendicular_midpoint_triangulation_method}. The
    position of the omnistereo sensor mounted on the quadrotor is annotated as frame $\fr{C}$ with
    respect to the scene's coordinates frame $\fr{S}$.
    {\bf(\subref{fig:point_cloud_synthetic_example_3D_view})} 3D visualization of the point cloud
    (the quadrotor with the omnistereo rig has been added for visualization only).
    {\bf (\subref{fig:point_cloud_synthetic_example_ortho_view})} Orthographic projection of the
    point cloud to the $\point{XY}$-plane of the visualization grid. 
    }
    \label{fig:point_cloud_synthetic_example}
\end{figure*}

By implementing the process described in \Fref{sec:3D_sensing},
 we attempt to reconstruct the
3D scene (as a point-cloud) obtained from the omnidirectional synthetic image given in
\Fref{fig:sim_rig_camera_view}, whose actual size is 1280x960 pixels.
The associated panoramic images, $\fr{\Xi_i}$, were obtained using function $\funcvv{h}_{\Xi_i}$
defined in \fref{eq:panorama_pixel_mapping_function} and are shown in \Fref{fig:panoramic_dense_stereo_matching}.
In fact, the point-cloud are generated using the dense stereo ``template matching'' technique
discussed in \Fref{sec:stereo_matching} and its resulting depth map is shown here, too. Pixel
correspondences $(\cfs{\pixelPoint{m}_1}{}{\Xi_1}, \cfs{\pixelPoint{m}_2}{}{\Xi_2})$ on the
panoramic representations are mapped via $\funcvv{h}_{\Xi_i}$ into their respective image positions
$\cfs{(\pixelPoint{m}_1, \pixelPoint{m}_2)}{}{I}$. Then, these are triangulated with
$\cfs{\funcvv{f}_{\point{P}_w}}{}{C}$ given in \fref{eq:function_triangulation_midpoint}, resulting in the set (cloud) of color 3D points $P_{\Delta}$ visualized in \Fref{fig:point_cloud_synthetic_example}.
Here, the synthetic scene (\Fref{fig:sim_rig_world_view}) is for a room
\SI{5.0}{\metre} wide (along its $\point{X}$-axis), \SI{8.0}{\metre} long
(along its $\point{Y}$-axis), and \SI{2.5}{\metre} high (along its $\point{Z}$-axis). 
With respect to the scene center of coordinates, $\fr{S}$, the
catadioptric omnistereo sensor, $\fr{C}$, is positioned at
$\cfs{\vect{t}}{C}{S}=\vectThreeColInline{1.60}{-2.85}{0.16}$ in meters.

\begin{figure*}
    \centering
    \includegraphics[width=1.0\textwidth,keepaspectratio]{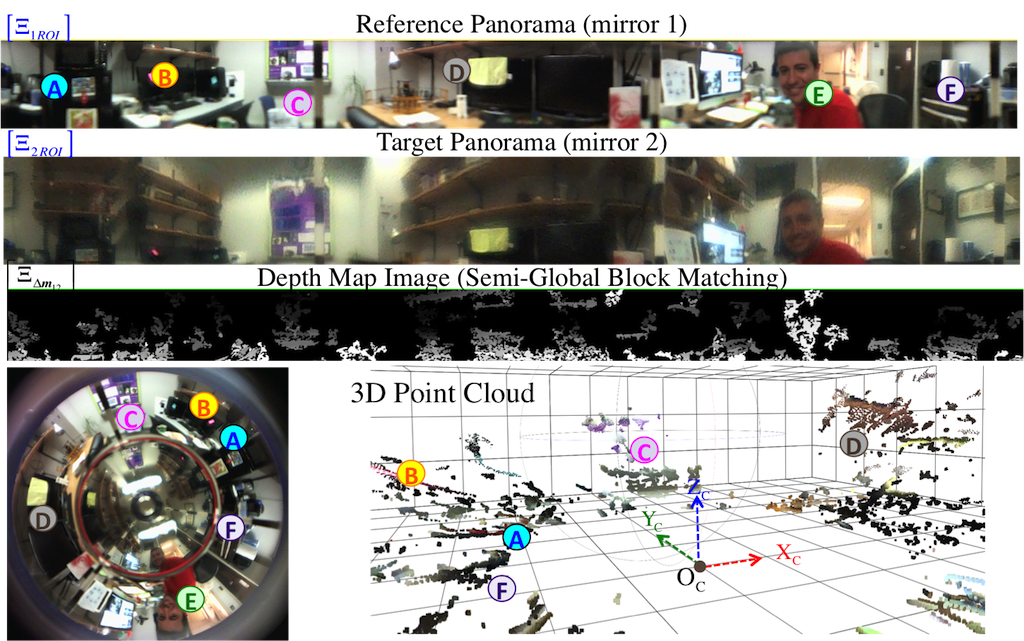}
    \caption[Real Omnistereo Experiment]{Real-life experiment using the
    \SI{37}{\milli\metre}-radius prototype and a single 2592x1944 pixels image
    (catadioptric view) from its color monocular camera. The rig was positioned in the middle of the
    room observed in \Fref{fig:74mm_prototype_on_pelican}. These panoramic images and depth map were
   obtained using the same procedure implemented for the synthetic model experiment. In order to better appreciate the resulting 3D point cloud, we have exagerated the point size and annotated several
    landmarks around the scene: 
    \\
    \mycirc{A} Appliances, \mycirc{B} left desktop, \mycirc{C} back wall,
    \mycirc{D} right desktop, \mycirc{E} person, \mycirc{F} toolbox. The grid size
    spacing is
    \SI{0.50}{\metre} in all directions.}
    \label{fig:real_omnistereo_experiment}
\end{figure*}

Finally, we present some preliminary results from a real
experiment using the prototype described in \Fref{sec:prototype_real} and shown in
\Fref{fig:74mm_prototype_on_pelican}. The panoramic images and point cloud shown in \Fref{fig:real_omnistereo_experiment}
are obtained by implementing the pertinent functions described throughout this manuscript and by
holding the SVP assumption of an ideal configuration. Merely, we provide these results as
proof of concept and we are aware that an adequate calibration procedure for the
proposed omnistereo sensor is desired.

\subsubsection{Sparse Triangulation Root-Mean-Square Error}
\label{sec:sparse_triangulation_rmse}

Due to the unstructured nature of the dense point clouds previously discussed, we proceed to
triangulate sets of sparse 3D points whose poses with respect to the omnistereo sensor
camera frame, $\fr{C}$, are known in advance. For various predetermined poses
$\cfs{\matx{T}_h}{G}{C}$, we synthetize a chessboard calibration frame $\fr{G}$ containing $m \times
n$ square cells.
Since the sensor is assumed to be rotationally symmetric,
it suffices to experiment with groups of 4 chessboard patterns situated at a given horizontal
range. A total of $4mn$ 3D points are available for each range group. Each corner point's position
$\cfs{\vect{p}_j}{}{C}$ is found with respect to $\fr{C}$ via the frame transformation
$\cfs{\vect{p}_j}{}{C}=\cfs{\matx{T}_h}{G}{C} \cfs{\vect{p}_j}{}{G}$ for all indices $j \in \{1, \ldots, 4mn\}$.

\begin{figure}
    \centering
    \includegraphics[width=\columnwidth,keepaspectratio]{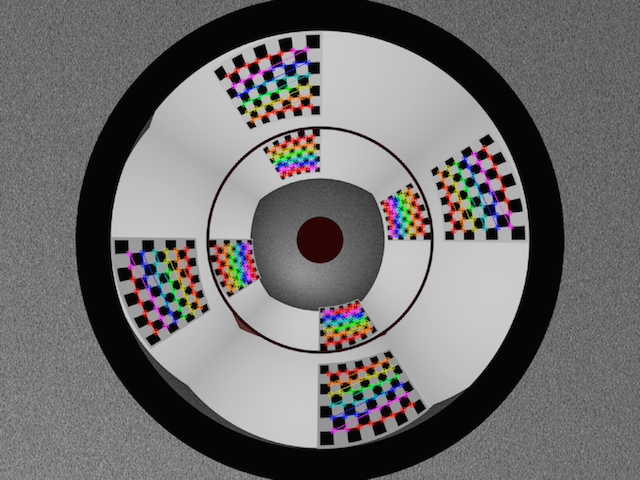}
    \caption[Real Omnistereo Experiment]{Example of sparse point correspondences detected from
    intersecting corners on chessboard patterns around the omnistereo sensor. The
    size of the rendered images for this experiment is 1280x960 pixels, and the cell size for this
    example's patterns is \SI{140x140}{\milli\metre}. The RMSE of this set of pattern points with
    origin points at $\cfs{\rho_{\point{O_G}}}{}{C} = \SI{2}{\metre}$ is approximately
    \SI{15}{\milli\metre} given in \Fref{tab:rmse_triangulation_experiment} and using a subpixel
    precision corner detector.
    }
    \label{fig:chessboard_patterns_omnidirectional_overlay}
\end{figure}

\Fref{fig:chessboard_patterns_omnidirectional_overlay} shows the set of detected corner points from
the group of patterns set to $\cfs{\rho_{\point{O_G}}}{}{C} = \SI{2}{\metre}$.
We adjust the pattern's cell sizes accordingly so its points
can be safely discerned by an automated corner detector \cite{Bradski2008}.
We systematically establish correspondences of pattern points on the omnidirectional image, and
proceed to triangulate with \fref{eq:function_triangulation_midpoint}.
For each range group of points, we compute
the root-mean-square of the 3D position errors (RMSE) between the observed (triangulated) points
$\cfs{\vect{\tilde{p}}_j}{}{C} \gets \funcvv{f}_{\point{P}_w}(\pixelPoint{\tilde{m}}_1,
\pixelPoint{\tilde{m}}_2)$ and the true (known) points $\cfs{\vect{p}_j}{}{C}$ that were used to
describe the ray-traced image.
\Fref{tab:rmse_triangulation_experiment} compiles the RMSE results and the standard deviation (SD)
for some group of patterns with origin, $\point{O_G}$, located at specified horizontal ranges
$\cfs{\rho_{\point{O_G}}}{}{C} \in [0.25,8.0]$ \si{\metre} away.

\begin{table}
    \caption{Results of RMSE from Triangulation Experiment}
    \label{tab:rmse_triangulation_experiment}
    \begin{center}
    \begin{tabular}
      {| S || S | S | }
        \hhline{|---|}
        {$\cfs{\rho_{\point{O_G}}}{}{C}$ [\si{\metre}]} & 
        {RMSE [\si{\milli\metre}]} &
        {SD [\si{\milli\metre}]}
        \\
        \hhline{|=#=|=|} 
        0.25  & 0.46 & 0.31
        \\
        0.50  & 1.20 & 0.71
        \\
        1.0  & 4.62 & 2.55
        \\ 
        2.0  & 14.85 & 9.06
        \\
        4.0  & 57.67 & 31.34
        \\
        8.0  & 219.09 & 129.92
        \\
        \hhline{|---|}
      \end{tabular}
      \end{center}
\end{table}

Finally, we notice that for all the 3D points in the patterns, we obtained an average error of $0.1
~\mathrm{px}$ with a standard deviation $\tilde{\sigma}_{px}= 0.05 ~\mathrm{px}$ for the subpixel
detection of corners on the image versus their theoretical values obtained from
$\funcvv{f}_{\varphi_i}$ defined in \eqref{eq:forward_projection_function}. This last experiment
helps us validate the pessimistic choice of $\sigma_{px}=1~\mathrm{px}$ for the discrete pixel space
in the triangulation uncertainty model proposed in \Fref{sec:triangulation_uncertainty_model}.


\subsection{Discussion and Future Work}
\label{sec:conclusion}

The portable aspect of the proposed omnistereo sensor is one of its greatest advantages (as
discussed in the introduction section). The total weight of the big rig using
\SI{37}{\milli\metre}-radius mirrors is about \SI{550}{\gram}, so it can be carried by the AscTec
Pelican quadrotor under its payload limitations of \SI{650}{\gram}. 
The mirror profiles have been optimized for the maximization of the stereo
baseline while obeying the various design constraints such as size and field of view.
Currently, the mirrors are custom-manufactured
out of brass using a CNC machine, and they seem to meet the performance specifications (e.g., FOV,
resolution).
We are hoping to reduce the system's weight dramatically by
employing aluminum or plastic-based mirrors.

In reality, it is almost impossible to assemble a perfect imaging system that fulfills the SVP
assumption and avoids the triangulation uncertainty studied in
\Fref{sec:triangulation_uncertainty_model} on top of the error already introduced by any feature  
matching technique.
The coaxial misalignment of the folded mirrors-camera system, defocus blur of the lens, and the
inauspicious glare from the support tube are all practical caveats we need to overcome for better 3D
sensing tasks. In fact, we avoided using the tube in our real-life experiment. 
The eminent need for an omnistereo calibration tool is part of our near future
endeavors.

The ongoing research is also focusing on the development of efficient software algorithms for
real-time 3D pose estimation and scene reconstruction via dense or sparse point clouds as to fit a particular
application.
Bear in mind that all the experimental results demonstrated in this manuscript rely upon a single
camera snapshot. We understand that the narrow vertical field-of-view where stereo vision operates
is a limiting factor for dense scene reconstruction from a single image. However, we believe that
robust feature tracking can lead to a robust pose estimation procedure by gathering key point
features all around the sensor. As in our past work \cite{Labutov2011}, the possibility of fusing
multiple modalities (e.g. stereo and optical-flow) is possible in order to resolve the scale-factor
problem inherent in the single, non-overlapping views.

In this work, we have validated the accuracy of synthetic experiments (in the ideal case), in
addition to an extensive study of the sensor's properties, such as its spatial resolution.
In order to validate the precision of the real sensor, we would require ground truth data obtained
from a relative position reference system, which we don't have at the moment. Our ultimate goal is
to apply the omnistereo sensor for autonomous flight using algorithms running on-board the
quadrotor's embedded computer.
3D simultaneous localization and mapping (SLAM) using the omnistereo vision system will be
integrated with our current autonomous navigation framework demonstrated in \cite{Valenti2014}.
At large, we have presented a thorough geometrical model and analytical solution to a useful
omnistereo sensor. We are willing to provide the corresponding software components upon request
until the final versions get implemented and released to the public.

\bibliographystyle{plain}
\bibliography{omni_MAV_references}

\end{document}